\pdfoutput=1
\documentclass[11pt]{article}
\usepackage{eamt24}
\usepackage{times}
\usepackage{url}
\usepackage{latexsym}
\usepackage[small,bf]{caption} % added MLF 20171211
\usepackage{soul}
\usepackage{tipa}
\usepackage{graphicx}
\usepackage{multirow}
\usepackage{amsmath}
\usepackage{subcaption}
\usepackage{url}
\usepackage{booktabs}
\usepackage{xcolor}

\usepackage[mathletters]{ucs}

\graphicspath{ {./figures/} }
\setuldepth{test}

\definecolor{hlcolor}{rgb}{.6,.99,.8}
\sethlcolor{hlcolor}

\title{Enhancing Gender-Inclusive Machine Translation \\ with Neomorphemes and Large Language Models}
\author{Andrea Piergentili,\textsuperscript{1,2}  Beatrice Savoldi,\textsuperscript{2}
 \textbf{Matteo Negri},\textsuperscript{2} \textbf{Luisa Bentivogli}\textsuperscript{2}\\
  \textsuperscript{1}University of Trento\\
  \textsuperscript{2}Fondazione Bruno Kessler\\
  {\tt \{apiergentili,bsavoldi,negri,bentivo\}@fbk.eu} \\}
\date{}

\begin{document}
 
\maketitle
\begin{abstract}

Machine translation (MT) models %have long been proven to 
are known to suffer from gender bias, especially when translating into languages with extensive
%extensively 
gendered morphology. %\bs{As such,} 
%Consequently,
Accordingly,
they still fall short in using gender-inclusive language, also representative of non-binary identities.
%suffer from gender bias, especially with grammatical gender languages, and have been recently been proven to fall short in using gender-inclusive language. %such as Italian. 
In this paper, we look at gender-inclusive neomorphemes, neologistic\footnote{
%We 
Following \cite{rose_variation_2023}, we refer to them as \textit{neologistic} because of their linguistically innovative nature.} elements that avoid binary gender markings
as an approach towards fairer MT.
%Along this line of research,
%Along this line of research, 
In this direction,
we
explore prompting 
techniques 
%to guide large language models (LLMs) 
%towards 
with large language models (LLMs) 
%\bs{for} translating 
to translate
from English into Italian using neomorphemes. 
So far, this area has been under-explored
%because of
due to
its novelty and the lack of publicly available evaluation resources.
%To fill this gap, we release
We fill this gap by releasing
\textsc{Neo-GATE},\footnote{Available at \url{https://huggingface.co/datasets/FBK-MT/Neo-GATE} under the CC-BY-4.0 licence.}
%\footnote{Upon paper acceptance, %the benchmark 
% data and code will be openly released% under \bs{the Apache 2.0 license.}
% .}
a 
resource
%corpus
designed to evaluate
%a benchmark for the evaluation of 
gender-inclusive 
en$\rightarrow$it translation 
%using 
%\bs{based on} 
with
%\neomorphemes. We experiment with different prompts and four LLMs of different families and sizes.
%
%
%\mn{neomorphemes. In addition,  through experiments with four LLMs of different families and sizes, as well as different prompt types, we identify strengths and weaknesses of each approach, laying the groundwork for future enhancements.}
neomorphemes. 
%Using
With
\textsc{Neo-GATE}, we assess 
four
%\mn{4}
LLMs of different families and sizes 
%as well as 
and different prompt %types, 
formats,
identifying strengths and weaknesses of each %approach 
on this novel task for MT.
% \mn{neomorphemes, and}
% experiment with %three prompting strategies and four incremental task demonstration 
% different prompts
% %several prompting settings
% and four LLMs of different families and sizes.
% %, three state-of-the-art open LLMs %-- Tower 7B instruct, LLama2 70B chat, and Mixtral 8x7B instruct -- and one commercial LLM. 
% %Our study shows that LLMs offer a viable path for this task when adequately prompted, but perform differently with different neomorpheme paradigms.
\end{abstract}

\section{Introduction}
%\looseness=-1 
%\bs{[Incipit molto diretto per i miei gusti ma puo' funzionare, pero' devi chiarire meglio perche' serve in MT e.g. i) per persone che non sono binary, e ii) per evitare gendered attribution quando non si sa il genere]}
Machine translation (MT) has been found to be susceptible to gender bias, i.e. the tendency to produce default masculine outputs %(e.g., Table \ref{tab:inclusive-translation-examples}, example A) 
or stereotypical gender associations \cite{saunders-etal-2020-neural,savoldi2021gender,Piazzolla_Savoldi_Bentivogli_2023} when gender information about human referents is absent. This is especially relevant when translating from notional gender languages like English, which express gender only through a limited set of elements (e.g., \textit{he/she} pronouns), into grammatical gender target languages, such as Italian, German, and Spanish, which mark gender extensively in their morphology (e.g., en: \textit{``My friends are rich''} $\rightarrow$ it: \textit{``\ul{I} mie\ul{i} amic\ul{i} sono ricch\ul{i}''} [M] vs \textit{``\ul{Le} mi\ul{e} amic\ul{he} sono ricch\ul{e}''} [F]).
%\mn{[METTEREI UNA TABELLA CON UN ESEMPIO E POI LO DISCUTEREI ESPLICITANDO LE IMPLICAZIONI, GLI HARMS: QUALI E PER CHI SONO HARMS. NELL'ESEMPIO METTEREI UNA FRASE INGLESE, LA SUA TRADUZIONE GENDERED, LA NEUTRALIZZAZIONE INDIRETTA ``PESANTE'' MA ACCETTABILE, LA NEUTRALIZZAZIONE INDIRETTA ``LEGGERA'', LA NEUTRALIZZAZIONE DIRETTA IN UN MODO, LA NEUTRALIZZAZIONE DIRETTA IN UN ALTRO MODO. Magari con caption ``ricca'' da cui emerga ad esempio che le neutralizzazioni dirette scelte sono 2 tra le tante possibili.]}
%
The consequences of this behavior are systematically harmful \cite{blodgett-etal-2020-language} to women, who risk being under-represented and stereotypically defined, and non-binary individuals, who are %inevitably 
erased from representation or misgendered within binary gender linguistic frameworks \cite{Misiek2020MisgenderedIT,dev-etal-2021-harms}.

In light of this, in this paper we look at neologistic solutions -- which are emerging from grassroots
efforts to make language more inclusive -- as a path towards gender-inclusive MT. Linguistic innovations such as neopronouns (e.g., en \textit{ze} instead of \textit{he/she}, sw \textit{hen} instead of \textit{han/hon}) and neomorphemes (e.g., it \textit{-\textschwa}/\textit{-\textrevepsilon}, es \textit{-e}/\textit{-es} in place of gendered inflectional morphemes) add new elements to the grammar and morphology to allow for the expression of non-binary gender identities or to convey gender neutrality \cite{bradley_singular_2019}.   
To date, the use of neologistic solutions is not systematized yet, with alternative paradigms coexisting and new ones %ever 
continuously
emerging. The choice and use of one paradigm of neologistic devices (e.g., the neopronouns \textit{xe/xem/xyr/xyrs/xemself} vs \textit{ze/zir/zir/zirs/zirself}, etc.) depends on individuals' identity and preferences in gender expression.

The use of neologistic devices in MT is still a largely unexplored research area, due to the novelty of this approach and to the %complete 
lack of dedicated
%training and evaluation 
resources, which in turn is complicated by the unfixed nature of these solutions. Ideally, gender-inclusive MT research should factor in the multiplicity of paradigms %which 
that
make up the landscape of neologistic devices \cite{lauscher-etal-2022-welcome}. However, the unavailability of evaluation and training resources which can be adjusted to any paradigm is a 
%critical bottleneck for both the investigation and the development
bottleneck for the investigation of gender-inclusive MT.
%
%Following recent research in performing GNT with LLMs \cite{savoldi2024prompt}, \bs{[sei un po' troppo autoreferenziale... ti rifai solo a noi come punti di attacco/motivazione principale per il lavoro. Qui puoi menzionare che gli LLM offrono una modalita' di adattabilita', oppure puoi menzionare previous attempts di Gosh and Caliskan... dire proviamo gli LLM perche' offrono possibilita' che l'MT non offre (vedi anche lavoro con i  neopronomi di Lauscher) e che queste possibilita' sono state anche verificate in paper eacl... \textbf{Ti consiglio anche questo paper anche per la motivazione sul forzare modelli a fare cose nuove non-standard: https://arxiv.org/pdf/2210.06245.pdf]}} \mn{[BEA HA RAGIONE; ALLARGA IL CAMPO AD ALTRI LAVORI PER DARE IDEA DI UNA LINEA DI LAVORO PIU' CONDIVISA CHE STA MATURANDO GRAZIE AI LLMs E NON UN CAPRICCIO DEGLI AUTORI DI EACL 24 :)]}
%
%\mn{[IL LAVORO PRECEDENTE APRE AGLI SVILUPPI PROPOSTI IN QUESTO PAPER, CHE COPRONO UN GAP TUTTAVIA ANCORA ESISTENTE. E' ORA DI FARLO MA NESSUNO L'HA FATTO COME NOI QUI. \textbf{BADA CHE QUI TI GIOCHI MOLTO SULLO SCORE DI NOVELTY.}]}
%
%Moreover
Also, neural MT 
%models 
%have 
has been proven to fail in handling neologistic gender-inclusive language \cite{lauscher-etal-2023-em}. Looking at other options, LLMs' ability to adapt to unseen tasks through in-context learning \cite{brown-2020-learners,min_rethinking_2022} offers a viable path toward gender-inclusive MT 
without the need for extensive training data.
%Therefore
Thus, in this work we investigate multilingual LLMs' ability to adapt to new, inclusive morphological paradigms in translation. %\mn{[PERCHE' LLMs CONSENTONO DI FARLO? PERCHE' HA SENSO PROVARCI? PERCHE' E' MEGLIO CHE PROVARCI CON TRADIZIONALE NMT? \textbf{BADA CHE QUI INIZI A GIOCARTI LO SCORE SU SOUNDNESS, almeno ad alto livello di approccio generale.}]}. 

%%\looseness=-1 
To this aim, we
%our contributions are: 
%\mn{we contribute with:}
\textit{i)}
%the release of
release
\textsc{Neo-GATE}, a benchmark 
%that allows to evaluate 
to evaluate
gender-inclusive en$\rightarrow$it translation with any of the ever-emerging neomorpheme paradigms;
\textit{ii)} 
%the exploration of 
explore
different prompting strategies for neologistic gender-inclusive MT, across three open and one commercial LLMs, and the two most popular Italian neomorpheme paradigms.

%\mn{[NON SO SE SOPRA HAI ELABORATO ABBASTANZA SUL FATTO CHE ESISTONO STRATEGIE DIVERSE. RIESCI QUI A DIRE EN PASSANT QUALI SONO, ALLUDENDO AL FATTO CHE VANNO PER LA MAGGIORE, O COMUNQUE RASSICURANDO IL LETTORE CHE LA TUA SCELTA E' SENSATA. NON LASCIARLO NEL DUBBIO CHE ABBIA SCELTO ARBITRARIAMENTE PARADIGMI MINORI O IN QUALCHE MODO PIU' FACILI DI ALTRI... \textbf{L'IMPACT NE SOFFRIREBBE}.]}.

\section{Background}
\label{sec:background}

\begin{table}[t]
\small
    \centering
    \begin{tabular}{ll}
        \textbf{EN}& I like being surrounded by my friends. \\
        \toprule
         \textbf{A}& Mi piace essere circondat\textbf{o} da\textbf{i} mie\textbf{i} amic\textbf{i}. \\
         \textbf{B}& \begin{tabular}[c]{@{}p{6.8cm}@{}} Mi piace avere \ul{persone amiche intorno a me}.\\
         \end{tabular}\\
          & \begin{tabular}[c]{@{}p{6.8cm}@{}} {[\scriptsize{en: I like to have people who are friends around me}]} \end{tabular} \\
         \textbf{C}&\begin{tabular}[t]{@{}p{6.8cm}@{}} Mi piace che \ul{intorno a me} siano presenti \ul{persone che considero mie amiche}. \end{tabular} \\ 
        & \begin{tabular}[c]{@{}p{6.8cm}@{}}{[\scriptsize{en: I like that there are people around me who I consider my friends}]} \end{tabular} \\
         \textbf{D}& Mi piace essere circondat\ul{*} da\ul{*} mie\ul{*} amic\ul{*}. \\
         \textbf{E}& Mi piace essere circondat\ul{\textschwa}~ da\ul{\textrevepsilon}~ mie\ul{\textrevepsilon}~ amic\ul{\textrevepsilon}. \\
         %\bottomrule
    \end{tabular}
    \caption{Examples of en$\rightarrow$it translations with no gender information in the source. Example A uses generic masculine formulations to refer to human beings (in bold), while the rest employ different gender-inclusive strategies (underlined). B and C use periphrases of different verbosity, while D and E employ different neomorpheme paradigms.}
    \label{tab:inclusive-translation-examples}
\end{table}

%%\looseness=-1 
%In light of this, and in an effort to move beyond the gender binary, 

%\pending{Following recent, unfolding social phenomena}
Following evolving
%social 
social and %language 
linguistic
phenomena
%trends}
%\bs{Within}
\cite{Gustafsson-2021,waldendorf_words_2023}, there has been a rising demand for the 
integration
%use 
of 
%a 
gender-inclusive language in natural language processing (NLP) %\bs{applications} 
technologies
to make them inclusive of all gender identities
%, beyond the gender binary 
\cite{dev-etal-2021-harms}.
In %translation technologies, 
MT,
gender-neutral translation (GNT) was recently proposed as a gender-fair approach to make %MT 
translation technologies
less biased and more inclusive \cite{piergentili-etal-2023-gender}. GNT consists in using gender-neutralization strategies, such as epicene %or impersonal 
formulations, (e.g., `persone' – en: \textit{people} – in examples B and C in Table \ref{tab:inclusive-translation-examples}), to avoid expressing the gender of human beings in the target language.
%
%\mn{[QUI FORSE DOVRESTI INTRODURRE PRIMA IL CONCETTO DI GNT \textbf{INDIRETTA}, MAGARI CON RIFERIMENTO ALL'ESEMPIO DI CUI SOPRA.]} However, \mn{indirect} 
%
However, this approach has considerable limitations: \textit{i)} it can result in verbose phrasings, 
%\mn{[``, come nell'esempio c in tabella 1, '']}
as in example C, which are only acceptable in specific contexts and textual domains, namely formal and institutional communication \cite{piergentili2023hi}; \textit{ii)} it is arguably impossible to translate some terms
%a \mn{an indirect}
by applying these strategies 
in grammatical gender languages like Italian (e.g., kinship terms, such as \textit{parent} → it \textit{genitore/genitrice}) \cite{motschenbacher2014grammatical}.
%\textit{iii)} though formally unmarked, gender-neutral expressions can still result in mental representations biased towards the masculine gender \cite{spinelli-2023}. 
%\mn{[ESEMPIO?]}.
Moreover, 
%gender-neutral language 
the use of circumlocutory language to avoid expressing gender
is regarded as a form of \textit{indirect} non-binary language \cite{attig-2020}, in that it conceals gender, while other, \textit{direct} solutions emphasize it.

%\bs{[Anche qui e' scritto bene, ma sei troppo verticale e autoriferito ai nostri lavori... io vedo la GNT come un lavoro di bacgkround da mettere insieme ad altri dopo come proposte alternative, non come il punto di discussione da cui partire.]}
Indeed, 
%contextually with the diffusion of gender-neutral language, 
%as a norm-abiding approach to gender-inclusivity in language \cite{piergentili-etal-2023-gender}, 
innovative alternatives have been proposed by queer communities as well. %In this section we discuss the principles and status of neologistic gender-inclusive language (Section \ref{sec:neologistic-gil}) and its adoption in NLP and MT (Section \ref{sec:related-work}).
%
%\subsection{Neologistic gender-inclusive language}
%\label{sec:neologistic-gil}
%%\looseness=-1 
Neologistic elements, such as neopronouns and neomorphemes, have emerged in notional gender languages, such as Swedish \cite{gustafsson2015hen} and English \cite{McGaughey2020-bm}, as well as in grammatical gender languages, such as Spanish \cite{Sarmiento2015desexualizacion}, French \cite{kaplan-2022-plurigrammars}, and German \cite{paolucci-etal-2023-gender}. %\bs{[non e' scontato per chiunque cosa siano notional e grammatical gender languages, non lo dici mai, devi srotolare all'inizio cosa implichi gramamtical gender language e qui puoi semplicemente parlare di gendered pronouns \mn{[+1]}]}
These devices aim to %\bs{complement [forte, sono risorse nuove per offire qualcosa che la traditional grammar non offre come possibilita' comuniticativa \mn{[+1]}]} 
enrich the %traditional grammar 
language with extra resources, which act as gender-neutral alternatives to gendered linguistic elements, and allow for a manifest inclusion of gender identities beyond the masculine-feminine binary \cite{bradley_singular_2019}. 
%\bs{e comunque, a parte il discorso standard/non standard, e' importante che in generale certe persone li usini proprio come identita' della propria vita... sia individui, sia per comunicazioni generiche.\mn{[Non so se e' quello che voleva dire Bea, ma mi fa pensare al fatto interessante che il neomorfema e' piu' esplicito dell'indiretta, e per questo potrebbe essere preferibile chi vuole affermare apertamente un'identita'. Quasi un atto politico? E' bello, e forse Bea dice proprio questo :)]}}
Individuals choose to use neologistic devices for themselves as they best fit their gender identity and as an open statement of it, rather than using gender-neutralization strategies, which would instead circumvent it \cite{gautam-2021-pronouns}.
Such innovative solutions are mostly used within LGBT+ communities, over informal channels.
%like social media \cite{Comandini_2021}. 
However, their use and acceptance are on the rise 
%\cite{waldendorf_words_2023}, especially among younger generations \cite{rose_variation_2023}.
\cite{waldendorf_words_2023,rose_variation_2023}.
While there is no \textit{one-fits-all} approach to gender-inclusive language \cite{lardelli_gender-fair_2023}, neologistic devices have naturally emerged as a response to the demand for a  
%direct and 
%direct, overarching 
%direct, all-encompassing 
direct solution that deserves attention.
%Thus, they should be taken into account as part of the gender-inclusive language landscape%, along with the gender-neutralization strategies mentioned above
% .

%%\looseness=-1 
In this work, we focus on the use of neomorphemes in 
%English~→~Italian 
en$\rightarrow$it
translation, a scenario in which gender-related 
ambiguities -- and, consequently, the need for gender-inclusive solutions -- are crucial. 
Indeed, %as a grammatical gender language, 
Italian is characterized by a pervasive gender-marking system, which assigns a gender to each noun and every word syntactically linked to it, including some verbal forms. %\bs{[p.s. serve prima questo, e non mi hai ancora dato manco mezzo esempio povera reader che sono :( \mn{[+100: tabella con esempio subito! E con tutti i concetti chiari fin dall'inizio!]} Mi parli di x strategie, lingue etc e non me fai vedere neanche una cosa esemplificativa.\mn{[siiiiiii!!!!]}]} 
Coherently, there have been several proposals of neomorpheme 
%paradigms for the Italian language, 
paradigms,
which currently coexist and are not yet ultimately codified \cite{sulis-and-gheno-2022-the-debate}. Such proposals promote the use of specific characters in place of gendered morphemes 
%(e.g., -o and -a, as in un\ul{o} scienziat\ul{o} [M], un\ul{a} scienziat\ul{a} [F] -- en: a scientist).
(e.g., masculine -o and feminine -a, as in ``un\ul{o} scienziat\ul{o}'' [M], ``un\ul{a} scienziat\ul{a}'' [F] -- en: a scientist).
%The characters proposed as 
The proposed
neomorphemes range from letters of the Latin alphabet (e.g., `u'$\rightarrow$un\ul{u} scienziat\ul{u}), to typographical symbols (e.g., `*'$\rightarrow$un\ul{*} scienziat\ul{*}), to letters of the international phonetic alphabet (IPA), 
%as is the case of 
like
the Schwa neomorpheme paradigm, which uses the IPA letter `\textschwa' for the singular number (un\ul{\textschwa} scienziat\ul{\textschwa}) and `\textrevepsilon' for the plural (alcun\ul{\textrevepsilon} scienziat\ul{\textrevepsilon} -- en: a few scientists) \cite{baiocco_italian_2023}.

\setlength{\tabcolsep}{1.5pt}
\begin{table*}[t]
\centering
 \small
\begin{tabular}{lll}

\toprule

% \multirow{1}{*}{\textbf{\textsc{GATE}}} & 
\begin{tabular}{l}\textbf{\textsc{GATE}} \end{tabular} & 
\begin{tabular}{ll}\textbf{Source}& \hspace{1pt} %At the meeting, 
The department chair said they might hire new professors \\

\textbf{Ref. Masc.}& \hspace{1pt} %Alla riunione, 
\hl{Il} \hl{direttore} del dipartimento ha detto che potrebbero assumere \hl{nuovi} \hl{professori} \\

\textbf{Ref. Fem.}& \begin{tabular}[c]{@{}p{12cm}@{}} \hspace{1pt} %Alla riunione, 
\hl{La} \hl{direttrice} del dipartimento ha detto che potrebbero assumere \hl{nuove} \hl{professoresse} \end{tabular}\end{tabular}\\

\midrule

\begin{tabular}{l}\textbf{\textsc{Neo-GATE}} \end{tabular} & \begin{tabular}{ll}
% \multirow{1}{*}{\textbf{\textsc{Neo-GATE}}} & \begin{tabular}{ll}
\textbf{Ref. tagged}& \begin{tabular}[c]{@{}p{12cm}@{}} %Alla riunione, 
\hl{\texttt{\textless{}DARTS\textgreater}} \hl{direttor\texttt{\textless{}ENDS\textgreater}} del dipartimento ha detto che potrebbero assumere \hl{nuov\texttt{\textless{}ENDP\textgreater}} \hl{professor\texttt{\textless{}ENDP\textgreater}}\end{tabular} \\

\textbf{Annotation}& \begin{tabular}[c]{@{}p{12cm}@{}} il la \texttt{\textless{}DARTS\textgreater}; direttore direttrice direttor\texttt{\textless{}ENDS\textgreater}; nuovi nuove nuov\texttt{\textless{}ENDP\textgreater}; professori professoresse professor\texttt{\textless{}ENDP\textgreater};\end{tabular}\end{tabular}\\

\midrule

% \multirow{1}{*}{\textbf{\textsc{Adapted *}}} & \begin{tabular}{ll}
% \textbf{Reference} &  \begin{tabular}[c]{@{}p{12cm}@{}} %Alla riunione, 
% \hl{L*} \hl{direttor*} del dipartimento ha detto che potrebbero assumere \hl{nuov*} \hl{professor*} \end{tabular}\\
% \textbf{Annotation} & \begin{tabular}[c]{@{}p{12cm}@{}} il la l*; direttore direttrice direttor*; nuovi nuove nuov*; professori professoresse professor*; \end{tabular}\end{tabular}\\

\begin{tabular}{l} {\textbf{\textsc{Neo-GATE}}} \\ {\textbf{\textsc{Adapted *}}}
\end{tabular} & \begin{tabular}{ll}
\textbf{Reference} &  \begin{tabular}[c]{@{}p{12cm}@{}} %Alla riunione, 
\hl{L*} \hl{direttor*} del dipartimento ha detto che potrebbero assumere \hl{nuov*} \hl{professor*} \end{tabular}\\
\textbf{Annotation} & \begin{tabular}[c]{@{}p{12cm}@{}} il la l*; direttore direttrice direttor*; nuovi nuove nuov*; professori professoresse professor*; \end{tabular}\end{tabular}\\

\midrule

% \multirow{1}{*}{\textbf{\textsc{Adapted \textschwa/\textrevepsilon}}} & \begin{tabular}{ll}
% \textbf{Reference} & \begin{tabular}[c]{@{}p{12cm}@{}} %Alla riunione, 
% \hl{L\textschwa} \hl{direttor\textschwa} del dipartimento ha detto che potrebbero assumere \hl{nuov\textrevepsilon} \hl{professor\textrevepsilon} \end{tabular} \\
% \textbf{Annotation} &  \begin{tabular}[c]{@{}p{12cm}@{}} il la l\textschwa; direttore direttrice direttor\textschwa; nuovi nuove nuov\textrevepsilon; professori professoresse professor\textrevepsilon; \end{tabular}\\

\begin{tabular}{l} {\textbf{\textsc{Neo-GATE}}} \\ {\textbf{\textsc{Adapted \textschwa/\textrevepsilon}}}
\end{tabular} & \begin{tabular}{ll}
\textbf{Reference} & \begin{tabular}[c]{@{}p{12cm}@{}} %Alla riunione, 
\hl{L\textschwa} \hl{direttor\textschwa} del dipartimento ha detto che potrebbero assumere \hl{nuov\textrevepsilon} \hl{professor\textrevepsilon} \end{tabular} \\
\textbf{Annotation} &  \begin{tabular}[c]{@{}p{12cm}@{}} il la l\textschwa; direttore direttrice direttor\textschwa; nuovi nuove nuov\textrevepsilon; professori professoresse professor\textrevepsilon; \end{tabular}\\

\end{tabular} \\
\bottomrule
\end{tabular}
\caption{%Examples of a \textsc{Neo-GATE} entry. The Adjusted references and annotations are the result of the automatic replacement of placeholder tags in the Tagged ref. and in the Annotation with the corresponding forms in the Asterisk and Schwa paradigms.
Examples of a single entry in GATE, \textsc{Neo-GATE}, and adapted to the two neomorpheme paradigms %we use
used
in our experiments 
%(see §\ref{sec:neomorphemes}). 
(§\ref{sec:neomorphemes}).
%The terms of interest in our evaluation, i.e. 
%those that express gender,
%the \pending{gendered} ones referring to human entities,
%are \hl{highlighted}.% \bs{\textbf{[xx]}}
The terms of interest for our evaluation are \hl{highlighted}.
}
\label{tab:testset-example}
\end{table*}

\paragraph{Gender Inclusivity in NLP} So far, research on gender-inclusive neologistic solutions in NLP has been mainly limited to first explorations in monolingual settings, and mostly confined to English neopronouns. In a pioneering effort, Lauscher et al.~\shortcite{lauscher-etal-2022-welcome} discussed the adoption of 
%neopronouns in NLP 
neopronouns
and formulated a list of desiderata 
% which 
to
model the use of pronouns in language technologies. They redefined pronouns as an \textit{open class}, i.e., %\bs{a category which should include standard gendered pronouns and any neopronoun paradigm each user may identify with.}
a %category 
class % set?
which is not fixed and allows for the inclusion of 
% unseen neopronoun paradigms each user may identify with. This is crucial when dealing with such novel devices, which are constantly evolving and emerging.
emerging neopronoun paradigms each user may identify with. This is crucial when dealing with such novel and constantly evolving devices.
%
%\bs{[per me che conosco a memoria questi lavori e' molto chiaro, ma secondo me devi portare per mano chi legge e spiegare cosa sia la questione open class e dire perche' ha valore questa cosa della open class nel tuo lavoro -- se davvero l'ha]}
%
%%\looseness=-1 
%Hossain et al.~\shortcite{hossain-etal-2023-misgendered} investigated the ability of seven LLMs to adapt to both the gender-neutral singular \textit{they} pronoun and several different neopronoun paradigms in a constrained decoding experiment, reporting poor performance by all models in the first case and even worse in the latter. %\mn{[IN COSA CONSISTE LA TUA NOVELTY RISPETTO A HOSSAIN? PERCHE' I RISULTATI SONO BASSI? C'E' UN LIMITE DI QUEL LAVORO CHE TU SUPERI?]}. 
%Similarly, in an open language generation experiment, Ovalle et al.~\shortcite{ovalle2023-towards-centering} prompted three decoder-only LLMs to use specific pronouns in association with a given referent, reporting generalized failures 
%and a prevalence of misgendering outputs 
%when eliciting the use of the singular \textit{they} or neopronouns. %\mn{[IN COSA CONSISTE LA TUA NOVELTY RISPETTO A OVALLE? PERCHE' I RISULTATI SONO BASSI? C'E' UN LIMITE DI QUEL LAVORO CHE TU SUPERI?]}
%In this wake, Brandl et al.~\shortcite{brandl-etal-2022-conservative} show how BERT-based models \cite{devlin-etal-2019-bert} produce high perplexity scores when processing English, Danish, and Swedish neopronouns in classification and coreference resolution tasks.
%%%%%%% SE NECESSARIO, DA QUI IN POI SI PUò TAGLIARE

In the context of generative tasks, several studies highlight the 
%shortcomings of LLMs in handling gender-inclusive neopronouns 
difficulty of LLMs %to handle 
in handling
neopronouns in zero-shot settings
\cite{brandl-etal-2022-conservative,hossain-etal-2023-misgendered,ovalle2023-towards-centering}.
Ovalle et al.~\shortcite{ovalle2023talking} identify byte pair encoding tokenization \cite{sennrich-etal-2016-neural} as a major cause of LLMs' shortcomings, coherently with Gaido et al.~\shortcite{gaido-etal-2021-split}, which observed the same phenomenon in gender bias investigation.
Tokenization, paired with the un-fixed nature of innovative gender-inclusive solutions, may represent a crucial problem for LLMs in correctly generating neomorphemes as well. Indeed, as mentioned above, the range of characters used as neomorphemes is wide and \textit{i)} not all characters are necessarily represented in the training data of LLMs; \textit{ii)} the use of different characters in place of more common gendered morphemes may result in different tokenizations for otherwise identical terms, which in turn could interfere with LLMs' ability to generate fluent text.
%%%%%% FINO A QUI

%\hl{[Spostare questa ref]} In a non-English experiment, Veloso et al.~\shortcite{veloso-etal-2023-rewriting} created a Portuguese gender-neutral rewriter which uses neopronouns and neomorphemes to convey gender-neutrality. %\mn{[IN COSA CONSISTE LA TUA NOVELTY RISPETTO A VELOSO? PERCHE' I RISULTATI SONO BASSI? C'E' UN LIMITE DI QUEL LAVORO CHE TU SUPERI?\textbf{REGOLA SEMPRE VALIDA: EVITA DI CITARE ALTRI LAVORI NEI RELATED WORKS SENZA POSIZIONARTI RISPETTO AD ESSI. SE QUESTI LAVORI SONO TUTTI FUORI DAL CONTESTO MT DEVI DIRLO CHIARAMENTE.}]}

%%\looseness=-1 
In MT 
%research specifically,
research, the sole experiment in developing 
%standard MT systems 
systems
partially compatible with neologistic devices is a proof-of-concept built by Saunders et al.~\shortcite{saunders-etal-2020-neural} in a gender bias mitigation experiment. They fine-tuned 
en$\rightarrow$de
%en~$\rightarrow$~de 
and 
%en~$\rightarrow$~es 
en$\rightarrow$es
MT models which use
%\ap{so as to generate}
placeholder tags in place of determiners and inflectional morphemes, 
%which can in turn 
to
be replaced with 
%the desired 
non-binary forms post-inference.
In a broader analysis of gender bias in LLMs, Vanmassenhove~\shortcite{vanmassenhove2024gender} reports that ChatGPT
%\footnote{\url{https://chat.openai.com/}} 
never produces gender-inclusive neomorphemes when translating ambiguous English sentences into Italian, %though 
although
without specifically prompting the model to do so. The sole analysis dedicated to the use of neologistic devices in MT is the %survey 
%\bs{reality-check}
%study
one
by Lauscher et al.~\shortcite{lauscher-etal-2023-em}, which shows how commercial 
%MT systems 
systems
fail to deal with English neopronouns, resulting in either misgendering or low-quality 
%output translations.
outputs.

A major bottleneck hindering the exploration of gender-inclusive neologistic devices in MT is the lack of publicly available evaluation resources. 
To bridge this gap, in the next section we introduce a dedicated resource: \textsc{Neo-GATE}.

%, a \bs{resource designed to benchmark the use of neologistic forms in automatic translation.}
%benchmark designed specifically for the evaluation of neologistic gender-inclusive MT.
%\bs{manca da qualche parte la questione dei LLM come potenziale approccio a superare limiti dell'MT. Potrebbe essere prima di questa parte ma non va dimenticata. E' assente sia in background che intro.} ANDREA: In realtà era nell'intro, nel penultimo paragrafo.

 %\bs{poi, vero che non riguarda l'uso di LLM e/o neologismi ma solo inclusivity in general in MT, pero' ci vorrebbero menzione a lavori come \cite{piergentili-etal-2023-gender} e \citep{savoldi2021gender} qui, magari anche solo anche an passant all'inizio dicendo... icnreasing efforts to make technologies more inclusive (cit) e poi avanti.}

\section{The \textsc{Neo-GATE} benchmark}
\label{sec:neo-gate}

%\bs{dividerei qui in sottosezioni dopo un cappello introduttivo al volo in cui dici in 2 parole cosa sia OPEN-gate e perche' scegliamo gate per crearlo. E poi farei sezione annotation e statistiche. E in chiusa dire che neomorfemi scegliamo di metterci qui. Vendilo bene suvvia povero dataset. \mn{[EFFETTIVAMENTE...BRAVA BEA.]}}

%%\looseness=-1 
\textsc{Neo-GATE} is designed to evaluate the use of neomorpheme paradigms in en$\rightarrow$it 
%translation tasks.
translation.
% en~$\rightarrow~$it translation%, thanks to a set of adaptable references and annotations. 
%en$\rightarrow$it.
%translation.
Following Lauscher et al.~\shortcite{lauscher-etal-2022-welcome}, and extending their desiderata to gender-inclusive translation,  
%in \textsc{Neo-GATE} we treat neomorphemes as an open class, i.e. embracing all existing and possible
we treat neomorphemes as an open class embracing %any 
all
possible neomorpheme paradigms. 
To this aim, we design \textsc{Neo-GATE} to be adjustable to any neomorpheme paradigm in Italian, thanks to a set of adaptable references and annotations.

%\looseness=-1 
\textsc{Neo-GATE} is built upon \textsc{GATE} \cite{rarrick2023gate}, a benchmark for the evaluation of gender bias in MT. 
 %\bs{an MT (binary) gender bias evaluation corpus, which consists of \textless{}\textit{source, M-reference, F-reference\textgreater{}} sentence triplets. 
 %As a matter of fact, GATE features gender ambiguous human entities (i.e. no explicit gender information are available) in each source English sentence, and for which two alternative masculine/feminine gendered reference translations are offered (SEE TABLE X).}
 %consisting of sentence-aligned triplets: \textit{i)} source English  }
 %corpus for MT with `gender-ambiguous' English source sentences and target language references which only differ in the gendered words.
In \textsc{GATE}, the gender of human entities is unknown, i.e. there are no elements providing gender information about human referents in the (English) source sentences. 
\textsc{GATE} also
%includes 
%\bs{provides two} 
provides
target language references which only differ in the feminine/masculine
%(binary) 
gendered words 
that refer to
%used to express 
human entities (see Table \ref{tab:testset-example}).
Since in our gender-inclusive translation task we envision the use of neomorphemes for human referents whose gender is unknown, GATE is an ideal candidate 
corpus as a basis for the creation of our resource.
\textsc{Neo-GATE} includes \textsc{GATE}'s test set entries,%\footnote{\textsc{GATE}'s en-it test set actually includes 843 entries. However, as two entries do not feature gender-marked terms in the references, we did not include them in \textsc{Neo-GATE}.}
\footnote{Except for two of GATE's entries, which do not feature gender-marked terms in the references.}
with the addition of %\bs{\textit{i)} 
references and 
(word-level) annotations 
based on
%tagged with 
a set of placeholder tags,
%\textbf{\bs{(see Appendix \ref{sec:appendix-tagset})}},
which can be automatically replaced with the desired forms.
%An example of a \textsc{Neo-GATE} entry is available in Table \ref{tab:testset-example}. 
The tagged references and the annotations are discussed in §\ref{sec:neogate-annotation}, while \textsc{Neo-GATE}'s evaluation metrics are described in §\ref{sec:evaluation}. %An example of a \textsc{Neo-GATE} entry is available in Table \ref{tab:testset-example}.}}

\subsection{Tagged references and annotations}
\label{sec:neogate-annotation}
For each entry in GATE's test set (see Table \ref{tab:testset-example} for an example), we want to create an additional reference translation featuring neomorphemes.
To this aim, for each gendered target word 
%in GATE, we 
we
% perform an annotation that replaces
replace
%we perform % word-level editing intended 
%to replace 
gendered morphemes and function words
%(articles, possessive adjectives, prepositions, etc.)
(articles, prepositions, etc.)
with placeholder tags.
% %
% Alongside \textsc{GATE}'s original masculine and feminine reference translations, \textsc{Neo-GATE} includes an additional \textbf{tagged reference}, which is analogous to the other two except it was edited so as to include placeholder tags in place of gendered morphemes and function words (articles, possessive adjectives, prepositions, etc.). 
%
%
%The placeholders serve the purpose of identifying words of interest for our task and making this reference adjustable to any neomorpheme paradigm by automatically replacing them with the desired forms.
The placeholders serve to identify words of interest for our task and make this reference adjustable to any neomorpheme paradigm by automatically replacing them with the desired forms.
% %%
The tagset was designed to cover all parts of the grammar which express grammatical gender, and accounts for distinct singular and plural forms 
% (e.g., the tags \texttt{<DARTS>} and \texttt{<DARTP>} stand for the singular definite article and the plural definite article, respectively.). 
(e.g., the tags \texttt{<DARTS>} and \texttt{<DARTP>} for the singular and plural definite articles respectively). 
This %is necessary to enable 
enables
the evaluation of 
% %% the neomorpheme paradigms which 
neomorpheme paradigms that
% %%account for the number grammatical category by using 
use different characters for the singular and the plural case, e.g., the `Schwa' paradigm mentioned in §\ref{sec:background}. 
While for content words we only replace the inflectional morpheme with a tag (either \texttt{<ENDS>} or \texttt{<ENDP>}), for function words we use placeholders that cover the whole word. We do so %for two reasons: 
because: \textit{i)} in Italian, some function words are not morphologically derived but paradigmatically opposed (e.g., the definite article singular masculine forms `il' and `lo' vs the feminine form `la'); \textit{ii)} as neomorpheme use is not yet settled, there are instances where competing forms exist for a single word and 
%these forms differ 
differ
in the root part (e.g., the forms `l\textrevepsilon' and `\textschwa' have been proposed\footnote{The first form was proposed in \url{https://italianoinclusivo.it/scrittura/}, and the second in \url{https://effequ.it/schwa/}} for the plural definite article). 
%In both cases,  it would be impossible to account for all possible forms with the sole inflectional placeholders, thus we replace all function words entirely with dedicated tags.
Since it would be impossible to account for all existing forms with the sole inflectional placeholders, we replace all function words entirely with dedicated 
%tags. Further details are reported in Appendix \ref{sec:function-words-annotation}.
tags (see Appendix \ref{sec:function-words-annotation} for further details).

We performed the same annotation
%editing 
on a subset of GATE's dev set as well, so as to have a pool of 
% blind 
exemplar sentences 
%to use 
for our experiments
%in the few-shot prompting experiments discussed below 
(see §\ref{sec:experimental-settings}).
%and to perform preliminary experiments blindly.
Table \ref{tab:tagset} in Appendix \ref{sec:appendix-tagset} 
%reports and describes 
describes
all the tags used in \textsc{Neo-GATE}, as well as the forms we used to replace them in our experiments. \textsc{Neo-Gate} statistics are reported in Table \ref{tab:stats}.

\begin{table}[h!]
    \centering
    \small
    \begin{tabular}{ccccccc}
    
         & \textbf{Entries} & \textbf{Tags} &  \textbf{Content}& \textbf{Function} & \textbf{Singular} & \textbf{Plural} \\
         \toprule
         \textbf{Test} & 841 & 2,479  & 1,539 & 940 & 1,316 & 1,163 \\
         \textbf{Dev} & 100 & 345 & 211 & 134 & 184 & 161 \\
        \bottomrule
    \end{tabular}
    \caption{Statistics of \textsc{Neo-GATE}'s test and dev sets.}
    \label{tab:stats}
\end{table}
% %%
% %%
% %%By creating a tagset mapping with the desired forms, it is possible to automatically replace the placeholder tags and use \textsc{Neo-GATE} to evaluate any neomorpheme paradigm.

%%\looseness=-1 
%After automatically identifying all words which should be annotated,\footnote{By extracting the differences in \textsc{GATE}'s original masculine and feminine references it is possible to automatically identify all words which should be edited.}
%the editing of the tagged reference was performed manually by a linguist, and %and validated through both automatic and manual checks. 
%%%%% SACRIFICABILE PER SPAZIO 
%Indeed, leveraging \textsc{GATE}'s original masculine and feminine references it is possible to automatically identify all words which should be annotated, so as to \textit{i)} annotate missing words and \textit{ii)} delete undue annotations. Moreover, by mapping the tagset with gendered forms, it is also possible to \textit{iii)} ensure that the correct tags have been used. 
%%%%% %%%%% %%%%% %%%%% %%%%%
%a second linguist\footnote{Both linguists are Italian speakers and authors of the paper.} performed an analogous annotation on a 15\% randomly selected subset of the corpus. As an indicator of inter-annotator agreement we compute\footnote{\url{scikit-learn.org/stable/modules/generated/sklearn.metrics.cohen_kappa_score}} Cohen’s kappa \cite{cohen-kappa}, which amounts to 0.94, indicating an almost perfect agreement \cite{landis1977}.

To ensure the quality of our resource, the %whole 
%corpus 
references were
%was 
manually annotated
%edited 
by a linguist 
%who followed 
following
dedicated guidelines.\footnote{The guidelines are available in \textsc{Neo-GATE}'s release page.} 
%Following
Using
the same guidelines, a second linguist\footnote{Both linguists are authors of the paper.} independently re-annotated a 15\% randomly selected subset of target language 
%sentences from the corpus. 
sentences.
%To measure inter-annotator agreement, we compute\footnote{\url{scikit-learn.org/stable/modules/generated/sklearn.metrics.cohen_kappa_score}} Cohen’s kappa \cite{cohen-kappa} on label assignment for the placeholder tags, which amounts to 0.94, indicating almost perfect agreement \cite{landis1977}.
Inter-annotator agreement computed with Cohen’s kappa \cite{cohen-kappa}\footnote{We use scikit-learn \cite{scikit-learn}.}
%\footnote{\url{scikit-learn.org/stable/modules/generated/sklearn.metrics.cohen_kappa_score}}
%\ap{using scikit-learn \cite{scikit-learn}}
on label assignment for the placeholder tags amounts to 0.94, indicating almost perfect agreement \cite{landis1977}. The few disagreements were overlooks and were thus reconciled.

%%\looseness=-1 
\textsc{Neo-GATE}'s set of annotated words %annotation 
is 
%\bs{then} 
automatically extracted by comparing the masculine, feminine, and tagged references.
%\bs{\textsc{Neo-GATE}}'s annotation 
It serves %the purpose of defining 
to define
%the words upon which the evaluation is based, i.e the words that will be searched for in models' outputs. 
the words upon which the evaluation is based.
It includes the three forms %of the terms %(e.g., `nuovi nuove nuov\texttt{<ENDP>}' in `Annotation', in Table \ref{tab:testset-example}) which are expected to be found in the outputs \mn{\textbf{[??????]}}. 
required for the evaluation, i.e the masculine and feminine forms, and the forms 
%indicated 
%annotated
with the placeholder tags, which are to be replaced with a 
%containing the 
neomorpheme (e.g., \textit{direttore}, \textit{direttrice}, and \textit{direttor*} in `\textsc{Neo-GATE Adapted *}', in Table \ref{tab:testset-example}).

\subsection{Evaluation metrics}
\label{sec:evaluation}

While holistic %reference-based 
metrics like BLEU \cite{papineni-2002-bleu} have been previously 
explored to inform the evaluation of
%used to assess 
gender bias in MT \cite{bentivogli-2020-gender,currey-etal-2022-mt},
%as done in \textsc{MuST-SHE} \cite{bentivogli-2020-gender} and \textsc{MT-GenEval} \cite{currey-etal-2022-mt}, 
these metrics are not designed to
%cannot 
provide %\mn{the fine-grained assessments we aim at DA ELABORARE.}
fine-grained assessments for specific linguistic phenomena. Rather, 
%holistic metrics 
they
offer a coarse-grained indication of overall translation quality, %\bs{which thus motivates} 
thus motivating
the use of dedicated metrics that allow for 
%more precise and focused evaluations, 
pinpointed evaluations,
isolating 
gender from other factors that could impact generic performance.
To this aim, we rely on %the
\textsc{Neo-GATE}'s %target 
annotations associated with
each source sentence (e.g. ``\textit{the department chair} said they might hire \textit{new professors}'' in Table \ref{tab:testset-example}). 
%Each
Every
annotation
comprises three forms for each gender-related word: masculine, feminine, and the form with neomorphemes (e.g. ``il la l*'', ``direttore direttrice direttor*'', ``nuovi nuove nuov*'', ``professori professoresse professor*''). 
In the description of our metrics, we refer to the total number of annotated triplets as `$annotations$' (4 triplets in our example).
Scores computation is carried out by scanning the models' output translations word by word, and checking whether such words match any of the three forms in the annotated triplets. 
%If a word match is found, we count it as `$matched$'.
Each matched word 
%contributes to increasing 
increases the `$matched$' count. If the matched form is the one with neomorphemes (e.g. ``direttor*''), we count it as `$correct$'.
To further monitor models' behavior %more broadly, 
we also count %how many 
the
%words that include a neomorpheme they generate, 
generated words that include a neomorpheme,
regardless of their presence in the annotations,
%. Each word that includes a neomorpheme increases the
in the
$found~neomorphemes$ tally. With these parameters, we %\bs{concretely} 
compute the following metrics.

\paragraph{Coverage (COV) and accuracy (ACC).}
As our primary evaluation method, we draw from the metrics defined by Gaido et al.~\shortcite{gaido-etal-2020-breeding} in the context of binary gender translation. Such a method first computes
%\ap{Following Gaido et al.~\shortcite{gaido-etal-2020-breeding}, we use gender-accuracy as our main method of evaluation.}
%We first define 
\textit{coverage} as the ratio of %annotations
annotated words %-- either masculine, feminine, or with neomorphemes -- 
$matched$ in the outputs over \textsc{Neo-GATE}'s %total number of 
$annotations$%, which we consider} data points: %$coverage  = \frac{matched}{total~data~points}$. 
:
$COV = \frac{matched}{annotations}$. 
This score serves two purposes: \textit{i)} it is indicative of the informativeness of the accuracy evaluation, as a low coverage indicates that the accuracy score described below is calculated over a relatively low number of %data points;
annotations;
%-- vice versa, a high coverage means that the %accuracy evaluation is more informative; 
\textit{ii)} %it is a generic indicator of the output quality, as a high coverage indicates that outputs contain either correctly generated translations with neomorphemes or wrong but fluent gendered translations. 
it can function as an indirect indicator of translation quality \cite{savoldi-etal-2022-morphosyntactic}, i.e. a higher coverage suggests that the model generates the expected target words.
% , as higher coverage suggests that the system generated the expected target word, 
% regardless of =
% gender \cite{savoldi-etal-2022-morphosyntactic}.

%As the evaluation of our task is similar %in principle 
%to gender bias evaluation, following Gaido et al.~\shortcite{gaido-etal-2020-breeding} we
On this basis, we then compute
%We define 
\textit{accuracy} as the proportion of 
% correct %terms generated by the model 
% %found 
% \ap{words, i.e. the ones that contain neomorphemes,}
%correctly generated neomorphemes
$correct$ neomorphemes generated by the model
over the total number of annotations %, either correct or gendered, 
$matched$ in the outputs: $ACC = \frac{correct}{matched}$. This score measures models' ability to correctly produce neomorphemes.
%Overall, the 

The
combination of 
these two metrics allows to distinguish between the generation of an annotated word (regardless of its gender) from its %actual 
gender realization 
%(feminine/masculine/neomorpheme),
(fem./mas./neom.),
thus 
ensuring pinpointed analyses.
%over the measurable data points.

%\paragraph{Over/under-generation}
%As in our task of gender-inclusive MT the models are required to use special characters, it is also important to evaluate whether they \textit{over-} or \textit{under-} generate them, regardless of how correctly they are using them. To do so we compute the \textit{over/under-generation} score relative to the total data points in the test set as $over{\text/}under\text{-}generation = \frac{-(total~d.~p.~-  all~matched~neomorphemes)}{total~data~points}$. This score complements the evaluation, as it can signal undesired model behaviors even despite good accuracy and coverage levels. Negative scores indicate that the model is generating fewer neomorphemes than it should, while positive scores mean that it is over-generating them, with values close to 0\% being ideal.
%This form of evaluation should only be performed when evaluating the use of neomorpheme paradigms which do not consist of regular characters of the Italian alphabet. Doing the opposite would result in an unreliable score, corrupted by unrelated but matched words. 

%%\looseness=-1 
\paragraph{Coverage-weighted accuracy (CWA).} 
% \noindent
% \textbf{Coverage-weighted accuracy.} 
%As a \textcolor{green}{general} indicator of models' performance in our task of gender-inclusive MT,
For a comprehensive view of models' overall 
%performance in gender-inclusive MT, 
performance,
%we compute a \textit{coverage-weighted accuracy}, which
CWA
takes into account both how accurately a model generates neomorphemes and the proportion of %\bs{the entire dataset} 
%the total %data points
%number of 
\textit{annotations}
covered by the evaluation:
% $coverage\text{-}weighted~accuracy = \frac{correct}{matched} *\frac{matched}{annotations}$.
\textit{CWA} $= \frac{correct}{matched} *\frac{matched}{annotations}$.
This score allows 
%for a comparison of the performance of different systems, 
%for comparing
for the comparison of
different systems, 
%and in different settings, 
for which both coverage and accuracy should be taken into account.
%%%%%%%% SACRIFICABILE PER SPAZIO %%%%%%%%
Indeed, a system's high accuracy may be the result of an evaluation %performed 
based
on a particularly small set of matched annotations, %corrupting 
impairing
the comparison with other systems' performance %if the evaluation is performed 
evaluated
on bigger portions of the corpus.
%%%%%%%%%%%%%%%%%%%%%%%%%%%%%%%%%%%%%%%%%%
While the other metrics %mainly 
serve to investigate each model's behavior, coverage-weighted accuracy allows for a fairer comparison %across 
of
different systems.

%%\looseness=-1 
\paragraph{Mis-generation (MIS).}
%To further investigate models' behavior in using neomorphemes in translation, we also consider the case where they generate neomorphemes inappropriately. Under realistic conditions, models could overgeneralize and generate neomorphemes also for unrelated words, compromising the intelligibility of the rest of the sentence. 
%
%
%To further investigate models' behavior in translating with neomorphemes, we also consider the case where they generate theminappropriately. 
We also consider the case where models generate neomorphemes 
%inappropriately. That is, for instance,
inappropriately, for instance
by applying the use of neomorphemes to %unrelated 
words %which 
that
do not refer to human entities (e.g., table$\rightarrow$tavol* instead of `tavolo').
Such scenario is  
% relevant to the extent that 
crucial, as
overgeneralizing the use of neomorphemes compromises the intelligibility of the %overall 
translation.
%This is the case of the over-generaton of spurious neomorphemes for unrelated words, which comes at the detriment  of translation quality.
%This occurs when models over-generate spurious neomorphemes for \mn{unrelated words}, which can negatively impact translation quality.
% \ap{For instance, this occurs when models over-generate spurious neomorphemes, including them 
% %in 
% \bs{for} unrelated words (e.g., adverbs, which do not express gender).}
Thus, to 
%\ap{intercept}
% this undesirable behaviour, 
we 
%estimate 
quantify
\textit{mis-generations}, i.e. the number of output words -- which are not annotated in \textsc{Neo-GATE} -- that feature neomorphemes.
%\ap{which are not part of the annotation and include neomorphemes.}
%
%\mn{MIS as the number of output words}
% -- which are not part of the
% annotation -- that are generated with neomorphemes. 
%\ap{which are not part of the annotation and include neomorphemes.}
%and do not match %\textsc{Neo-GATE}'s annotation. 
% \ap{Consistently with the other metrics, we compute it as a ratio over the total data points:
% $mis\text{-}generation = \frac{matched~neomorphemes - correct}{total~data~points}$.}
%\mn{Mis-generation is computed as the ratio of THE DIFFERENCE BETWEEN A e B,OVER...correctly matched neomorphemes over the total data points:}
Accordingly, 
%mis-generation 
%\mn{MIS} is computed as
% \ap{Mis-generation is computed as %the ratio of 
% the difference between the $found~neomorphemes$ and the $correct$ neomorphemes, over the $annotations$.}
%
%
%
% $mis\text{-}generation = \frac{found~neomorphemes~-~correct}{annotations}$.
$MIS = \frac{found~neomorphemes~-~correct}{annotations}$.
This score complements 
%\bs{the view offered in terms of accuracy and coverage.}
% the evaluation, as it can signal undesired model behaviors (e.g., over-generation, or the undue use of neomorphemes in general) even despite good accuracy and coverage.
the evaluation,  as it can signal
%\mn{by signalling} 
undesired %model 
behaviors even despite good accuracy and coverage.

\section{Experimental settings}
\label{sec:experimental-settings}

\setlength{\tabcolsep}{2.5pt}
\begin{table}[ht!]
\small
    \centering
    \begin{tabular}{lcccccc}
    
&  \textbf{\textsc{BLEU}}& \textbf{\textsc{chrF}}& \textbf{\textsc{TER} $\downarrow$}& \textbf{\textsc{BERTSc.}}& \textbf{\textsc{COMET}}\\
         \toprule
         %\textbf{opus-mt-en-it}
          \textbf{\textsc{opus-mt}}
         & 27.53& 57.61& 58.95& 87.42&  82.68 \\
         \midrule
         \textbf{GPT-4}& \textbf{\underline{32.34}}& \textbf{\underline{61.11}}& \textbf{\underline{54.87}}& \textbf{\underline{88.76}}& \textbf{\underline{87.05}} \\
         \textbf{Tower}& \underline{30.88} & \underline{59.41} & \underline{56.96} & \underline{88.17} &  \underline{86.21} \\
         \textbf{Mixtral}& \underline{29.63}& \underline{58.68}& 59.35& \underline{87.81}&  \underline{86.11} \\
         \textbf{LLama 2}& 26.28& 55.92& 61.98& 87.02& \underline{84.23} \\
         
         \bottomrule
    \end{tabular}
    \caption{Translation quality results. Best scores are in \textbf{bold}.
    Cases where LLMs outperform the MT model are \underline{underlined}.
    }
    \label{tab:translation-quality-results}
\end{table}

\subsection{Models}
\label{sec:models}
%%\looseness=-1 
%We experiment with three open, decoder only LLMs: TowerInstruct-7B-v0.1,\footnote{\url{https://huggingface.co/Unbabel/TowerInstruct-7B-v0.1}} Mixtral-8x7B-Instruct-v0.1 \cite{jiang2024mixtral}, and LLama 2 70B chat \cite{touvron2023LLama}. 
We experiment with three open, decoder only LLMs. 
%One model, 
TowerInstruct-7B-v0.1,\footnote{\url{https://huggingface.co/Unbabel/TowerInstruct-7B-v0.1}} is fine-tuned for MT, whereas the other two
%, 
-- Mixtral-8x7B-Instruct-v0.1 \cite{jiang2024mixtral} and LLama 2 70B chat \cite{touvron2023LLama} -- are
not specialized for MT.
%general-purpose LMs.}
%For comparison, 
We
%We 
also include
the 
%proprietary, commercial model GPT-4
commercial model GPT-4
\cite{openai2024gpt4},\footnote{Model \texttt{gpt-4-0125-preview}} which proved to perform well in %previous
gender-inclusive MT experiments \cite{savoldi2024prompt}. We use 
the models' 
default settings, except for the temperature parameter, which we set to 0
%.00 
following Peng et al.~\shortcite{peng-etal-2023-towards}. 
We do not include neural MT models as no model currently supports neomorphemes and no dedicated training or fine-tuning data is available.

\begin{table*}[ht]
\centering
\small
\begin{tabular}{lll}
\textbf{\textsc{Prompt}}& \textbf{\textsc{Role}}& \textbf{\textsc{Example}} \\
\toprule
\textbf{Zero-shot}& \texttt{user}& \begin{tabular}[c]{@{}p{12.4cm}@{}}Translate the following English sentence into Italian using the neomorpheme `*'. To do so, the neomorpheme `*' should be used as a substitute for masculine and feminine morphemes in words that refer to human beings.\\ {[}English{]} \texttt{\textless{}\{input sentence\}\textgreater{}} \\
{[}Italian{]}\end{tabular}   \\
\midrule
\multirow{3}{*}{\textbf{Direct}}& \texttt{user}& \begin{tabular}[c]{@{}p{12.4cm}@{}}{[}English{]} \textless{}I never buy flowers for my friends.\textgreater{}\\ {[}Italian{]}\end{tabular} \\
\cmidrule{2-3}
& \texttt{assistant}&\textless{}Non compro mai fiori per \hl{l*} \hl{mi*} \hl{amic*}.\textgreater{}\\
 \midrule
\multirow{5}{*}{\textbf{Binary}}& \texttt{user}& \begin{tabular}[c]{@{}p{12.4cm}@{}}{[}English{]} \textless{}I never buy flowers for my friends.\textgreater{}\\ {[}Italian, gendered{]}\end{tabular} \\
\cmidrule{2-3}
& \texttt{assistant}& \begin{tabular}[c]{@{}p{12.4cm}@{}}\textless{}Non compro mai fiori per \hl{i} \hl{miei} \hl{amici}.\textgreater{}\\
  {[}Italian, neomorpheme{]} \textless{}Non compro mai fiori per \hl{l*} \hl{mi*} \hl{amic*}.\textgreater{}\end{tabular} \\
  \midrule
\multirow{4}{*}{\textbf{Ternary}}& \texttt{user}& \begin{tabular}[c]{@{}l@{}}{[}English{]} \textless{}I never buy flowers for my friends.\textgreater{}\\ {[}Italian, masculine{]}\end{tabular} \\
\cmidrule{2-3}
& \texttt{assistant}& \begin{tabular}[c]{@{}l@{}}\textless{}Non compro mai fiori per \hl{i} \hl{miei} \hl{amici}.\textgreater{}\\ {[}Italian, feminine{]} \textless{}Non compro mai fiori per \hl{le} \hl{mie} \hl{amiche}.\textgreater{}\\ {[}Italian, neomorpheme{]} \textless{}Non compro mai fiori per \hl{l*} \hl{mi*} \hl{amic*}.\textgreater{}\end{tabular} \\   
 \bottomrule
\end{tabular}
\caption{Examples of all the prompts used in our experiments. The few-shots prompt examples include the Asterisk neomorpheme. Words expressing gender are \hl{highlighted}. %\textbf{\bs{QUALCOSA DI STRANO IN STA TABELLA}}
}
\label{tab:prompt_examples}
\end{table*}

To ensure the suitability of the selected LLMs for translation-related tasks, we preemptively test their generic 
en$\rightarrow$it translation performance on the \textsc{FLORES 101} benchmark \cite{goyal-etal-2022-flores}.
%To this aim, we use \textsc{FLORES 101} \cite{goyal-etal-2022-flores} as a benchmark for general translation quality and 
We prompt the models to translate with a few-shot prompt (see Appendix \ref{sec:appendix-mt-prompt}). In these experiments, we include \texttt{opus-mt-en-it},\footnote{\url{https://huggingface.co/Helsinki-NLP/opus-mt-en-it}} a state-of-the-art neural MT model, as a reference system for translation quality. For general MT evaluation, we use
%we use five %automatic metrics: 
BLEU% \cite{papineni-2002-bleu}
, chrF \cite{popovic2015chrf}, TER \cite{snover-etal-2006-study}, %METEOR \cite{banerjee2005meteor}, 
BERTScore \cite{zhang2019bertscore}, 
and COMET \cite{rei2020comet}. 
Using %this set of 
%theese
these
metrics allows for 
%an evaluation and comparison 
a comparative evaluation
of %models' 
translation performance based on different aspects, namely the surface similarity to human-made reference translations (BLEU, chrF, TER), and the semantic adherence to those references (BERTScore) and to the source (COMET).
The results 
%are reported in  Table \ref{tab:translation-quality-results}, and show that 
in Table \ref{tab:translation-quality-results} show that
the LLMs 
perform very well in MT, often outperforming the SOTA MT system in this setting.

% \setlength{\tabcolsep}{2.5pt}
% \begin{table}[t]
% \small
%     \centering
%     \begin{tabular}{lcccccc}
    
% &  \textbf{\textsc{BLEU}}& \textbf{\textsc{chrF}}& \textbf{\textsc{TER} $\downarrow$}& \textbf{\textsc{BERTSc.}}& \textbf{\textsc{COMET}}\\
%          \toprule
%          %\textbf{opus-mt-en-it}
%           \textbf{\textsc{opus-mt}}
%          & 27.53& 57.61& 58.95& 87.42&  82.68 \\
%          \midrule
%          \textbf{GPT-4}& \textbf{\underline{32.34}}& \textbf{\underline{61.11}}& \textbf{\underline{54.87}}& \textbf{\underline{88.76}}& \textbf{\underline{87.05}} \\
%          \textbf{Tower}& \underline{30.88} & \underline{59.41} & \underline{56.96} & \underline{88.17} &  \underline{86.21} \\
%          \textbf{Mixtral}& \underline{29.63}& \underline{58.68}& 59.35& \underline{87.81}&  \underline{86.11} \\
%          \textbf{LLama 2}& 26.28& 55.92& 61.98& 87.02& \underline{84.23} \\
         
%          \bottomrule
%     \end{tabular}
%     \caption{Translation quality results. Best scores are in \textbf{bold}.
%     Cases where LLMs outperform the MT model are \underline{underlined}.
%     }
%     \label{tab:translation-quality-results}
% \end{table}

\subsection{Neomorphemes}
\label{sec:neomorphemes}
%%\looseness=-1 
%In our experiments, we
We
focus on the two most popular %and productive \mn{\textbf{[??????]}}  
Italian neomorpheme paradigms \cite{Comandini_2021}: \textit{i)} the \textit{Asterisk}, which uses the symbol `*' as a graphemic device in place of regular inflectional morphemes \cite{haralambous}; 
% \textit{ii)} the \textit{Schwa} paradigm, 
\textit{ii)} the \textit{Schwa},
which features both a singular form, for which the character `\textschwa' is used, and a plural 
form, represented with 
%\mn{form using}
the character `\textrevepsilon' \cite{baiocco_italian_2023}.

%\looseness=-1 
%%% ORIG ANDREA
%We created a tagset mapping for the Asterisk paradigm and one for the Schwa paradigm. Each mapping associates the tags used in the tagged references to the desired form for the specific paradigm. The mappings are reported in Appendix \ref{sec:appendix-tagset}.
%
%
%\mn{For each paradigm we created a tagset mapping, each associating the tags used in the tagged references to the desired form for the specific paradigm.}
\looseness=-1
For each paradigm, we create a tagset mapping (see Appendix \ref{sec:appendix-tagset}) that associates the tags used in the tagged references with the desired form for that specific paradigm.
As no complete %and shared 
codification of the use and the orthography of neomorphemes in Italian is available \cite{thornton_genere_2020}, we referenced established 
%informative resources 
resources
such as the website \textit{Italiano Inclusivo},\footnote{\url{https://italianoinclusivo.it/scrittura/}} and examples %reported 
found
in scientific literature, such as Rosola et al.~\shortcite{rosola2023beyond}. As these sources do not cover the whole set of possibly gendered elements in the grammar, we derived the missing forms by analogy from elements of the same class. For example, 
since 
%as
none of these sources describes the full set of articulated prepositions, which 
%that
express gender in Italian,
%thus we 
we
used the given examples as a model for the rest of the class% (see Appendix \ref{sec:appendix-tagset})\mn{\textbf{[??????]}}.
.

\subsection{Prompts} 
\label{sec:prompts}
% We experiment with four prompt formats.
% %in one \bs{in} zero-shot and three few-shots settings.
% Namely, one
% %The 
% zero-shot prompt that only consists 
% in 
% %\bs{of} 
% a verbalized instruction, whereas the few-shot prompts include the same verbalized instruction along with task demonstrations with different formats (see also Table \ref{tab:prompt_examples}):
%
% \mn{We experiment with four prompting strategies: one zero-shot prompt consisting solely of a verbalized instruction, and three few-shot prompts containing the same verbalized instruction along with task demonstrations in various formats. Specifically, as illustrated by the examples in Table \ref{tab:prompt_examples}:}
% \mn{We experiment with four zero/few-shot prompting strategies illustrated by the examples in Table \ref{tab:prompt_examples}:}

% We experiment with four prompt formats, also illustrated by the examples in Table
We experiment with one zero-shot and three few-shot formats, illustrated by the examples in Table \ref{tab:prompt_examples}.
The few-shot prompts follow the format used in Sánchez et al.~\shortcite{sánchez2023genderspecific}, which was found to be useful for controlling gender expression in translation. %and emerged as the best performing and the least computationally expensive one in Savoldi et al.~\shortcite{savoldi2024prompt}. 
%The few-shot prompts 
We instantiate different conceptualizations of the task, ranging from a simple pairing of %English 
source sentences directly with gender-inclusive %Italian 
translations, to a ternary opposition of masculine, feminine, and gender-inclusive translations:

% \begin{itemize}
%     \item \textbf{Zero-Shot:} a verbalized description of the task is provided without any demonstration.
%     \item \textbf{Direct:} the same verbalized instruction is provided along with demonstrations that include the English source sentence and the gender-inclusive Italian translation.
%     \item \textbf{Binary:} an intermediate gendered (masculine) Italian translation is  also included in the format. This format follows the one used in Savoldi et al.~\shortcite{savoldi2024prompt}, which frames the task as a double output translation. The models are asked to produce a gendered translation first and then a second one with neomorphemes, which should be identical to the first except for the words expressing gender.
%     \item \textbf{Ternary:} two intermediate gendered translations  (one masculine, one feminine) are included. The rationale for this format is that, by instantiating a ternary opposition, the models may better identify parts of the target language sentences that should be identical among the three translations and the parts that should differ, i.e. those expressing gender. Framing the task as a triple output translation could help the models infer that the gender expressed in the third translation should be something other than masculine and feminine.
% \end{itemize}

%%%%%%%%%%%%%%%%%%%%%%%%%%
\noindent
${\Diamond}$\textbf{Zero-Shot:} a verbalized description of the 
%task, 
task is provided
without any demonstration. 

% The same instruction is also included %in the first \texttt{user} message of 
% \ap{at the beginning of}
% the few-shot prompts.

% \noindent${\Diamond}$\textbf{Direct:} the demonstrations include the English source sentence and the gender-inclusive Italian translation.

\noindent${\Diamond}$\textbf{Direct:} the same verbalized instruction is provided along with demonstrations %including 
that include
the English source sentence and the gender-inclusive Italian translation.
% \paragraph{Direct:} the same verbalized instruction is provided along with demonstrations %including 
% that include
% the English source sentence and the gender-inclusive Italian translation.

%\looseness=-1 
\noindent${\Diamond}$\textbf{Binary:} an intermediate gendered 
%-- masculine -- 
(masculine)
Italian translation is 
also included in the format.
%\mn{included in the demonstration.}
This format follows the one used in Savoldi et al.~\shortcite{savoldi2024prompt}, which frames the task as a double output translation. The models are asked to produce a gendered translation first and then a second one with neomorphemes, which should be identical to the first except for the words 
%which express gender.
expressing gender.

%\looseness=-1 
\noindent ${\Diamond}$\textbf{Ternary:} two intermediate gendered translations 
%-- one masculine, one feminine -- 
(one masculine, one feminine)
%are included in the format. 
are included.
The rationale for this format is that, by instantiating a ternary opposition, the models may better identify parts of the target language sentences 
%which
that
should be identical among the three translations and 
the parts
%which
that
should differ, i.e. those 
%that express 
expressing
gender. Framing the task as a triple output translation could help the models infer that the gender expressed in the 
%final translation
%\mn{translation}
third translation
should be something other than 
%the masculine and the feminine.
masculine or feminine.
%%%%%%%%%%%%%%%%%%%%%%%%%%

%\looseness=-1 
% The few-shot prompts follow the format used in Sánchez et al.~\shortcite{sánchez2023genderspecific}, which was found to be useful for controlling gender expression in translation. %and emerged as the best performing and the least computationally expensive one in Savoldi et al.~\shortcite{savoldi2024prompt}. 
% %The few-shot prompts 
% We instantiate different conceptualizations of the task, ranging from a simple pairing of %English 
% source sentences directly with gender-inclusive %Italian 
% translations, to a ternary opposition of masculine, feminine, and gender-inclusive translations.
%In all formats, we 
We %include 
enclose
the exemplar sentences 
%and the input source sentence 
in angle brackets \texttt{<>}. Models are expected to reproduce this structure, thus facilitating the extraction of the final translation from %the whole 
% their
the
output in postprocessing.
%Examples of each prompt format are shown in Table \ref{tab:prompt_examples}.

%%%% ORIG ANDREA
%All four models expect prompts in a `chat' format, i.e. structured as an alternation of \texttt{user} messages, which provide the input to the model, and \texttt{assistant} messages, which represent the model's desired output.\footnote{\url{https://huggingface.co/docs/transformers/main/en/chat_templating}} Thus, we format the few-shots prompts according to this structure, whereas for the zero-shot ones we only input a single \texttt{user} message.
%
%%%%% REVISED MN
% \mn{All four models expect prompts in a `chat' format, structured as an alternation of \texttt{user} messages (providing input to the model) and \texttt{assistant} messages (representing the model's desired output).\footnote{\url{https://huggingface.co/docs/transformers/main/en/chat_templating}} We thus format the few-shot prompts according to this structure, while  for the zero-shot ones we only input a single \texttt{user} message.}
%%%% SHORTER MN
All four models expect prompts in a `chat' format, with \texttt{user} messages providing input and \texttt{assistant} messages representing the model's desired output.\footnote{\url{https://huggingface.co/docs/transformers/main/en/chat_templating}} For the few-shot prompts we adhere to this structure, whereas for the zero-shot prompts, we only provide a single \texttt{user} message.

\paragraph{Demonstrations} In the 
%few-shots 
few-shot
settings (i.e. Direct, Binary, Ternary) we included \textbf{1}, \textbf{4}, and \textbf{8} task demonstrations in the prompts. The extremes were chosen as the minimum necessary to elicit 
in-context learning 
%(ICL)
(1) and a compromise between  
a high number
%\mn{the number}
of demonstrations and the computational cost of inference (8). The exemplar sentences were selected from \textsc{Neo-GATE}'s dev set (§\ref{sec:neogate-annotation}). 
%The exemplars were selected so as to be representative of the dev set's average tag \textit{density}, i.e the amount of tags in each reference, and to provide a good balance of singular and plural tags. The prompts were then formatted with each paradigm's tagset mapping before prompting the model.
The exemplars were chosen so as to represent the average tag \textit{density} of the dev set, i.e., the number of tags in each reference, and to offer a balanced mix of singular and plural tags. The prompts were then formatted using each paradigm's tagset mapping before presenting them to the model.

\section{Results and discussion}
\label{sec:results}
% %In this Section
% Here we discuss the results of our experiments in zero-shot (§\ref{sec:results-zero-shot}) and few-shots settings (§\ref{sec:results-few-shots}).

%Table \ref{tab:results-zero-shot} reports the accuracy, coverage, and over/under-generation scores for the zero-shot experiments, while for few-shots experiments we report the accuracy and coverage scores in Table \ref{tab:results} and the over/under-generation scores in Table \ref{tab:results-over}. All the coverage-weighted accuracy scores are visualized in Figure \ref{fig:coverage-weighted-accuracy}, and the detailed results are reported in Table \ref{tab:detailed-cov-weigh-acc}. 

%%%%%%%%%%%% TABELLA zero-shot %%%%%%%%%%%%%%%%
\setlength{\tabcolsep}{4pt}
\begin{table}[ht]
\centering
\small
\begin{tabular}{lcccc}
\toprule
   %\multicolumn{5}{c}{\textbf{Asterisk}} \\
   %\midrule
 \textbf{\textsc{Asterisk}} & \textbf{COV~$\uparrow$}& \textbf{ACC~$\uparrow$}& \textbf{CWA~$\uparrow$} & \textbf{MIS~$\downarrow$} \\
\midrule
\textbf{GPT-4} & 57.08& \textbf{74.63}&  \textbf{42.60}  & 45.78   \\
%\midrule
\textbf{Tower}  & \textbf{77.57} & 0.00 & 0.00 & \textbf{0.00}  \\
\textbf{Mixtral}&  35.22& 37.92& 13.35 & 52.20  \\
\textbf{LLama 2}&  56.72& 0.57 & 0.32 & 16.70 \\
\midrule
%\multicolumn{5}{l}{\textbf{Schwa}} \\
%\midrule
\textbf{\textsc{Schwa}} & \textbf{COV~$\uparrow$}& \textbf{ACC~$\uparrow$}& \textbf{CWA~$\uparrow$} & \textbf{MIS~$\downarrow$} \\
\midrule
\textbf{GPT-4} & 46.91& \textbf{60.19}& \textbf{28.24} & 72.77 \\
\textbf{Tower} &\textbf{77.25}& 0.00 & 0.00 & \textbf{0.00} \\
\textbf{Mixtral} & 30.05& 27.79& 8.35 & 61.44 \\
\textbf{LLama 2}&  57.60 & 0.35 & 0.20 & 12.79 \\
\bottomrule
\end{tabular}
\caption{Zero-shot setting results.
We report the coverage (COV), accuracy (ACC), coverage-weighted accuracy (CWA), and mis-generation (MIS) scores.}
\label{tab:results-zero-shot}
\end{table}

\begin{figure*}[ht]
    \centering
\begin{subfigure}{0.49\textwidth}
    \centering
    \includegraphics[width=.99\linewidth]{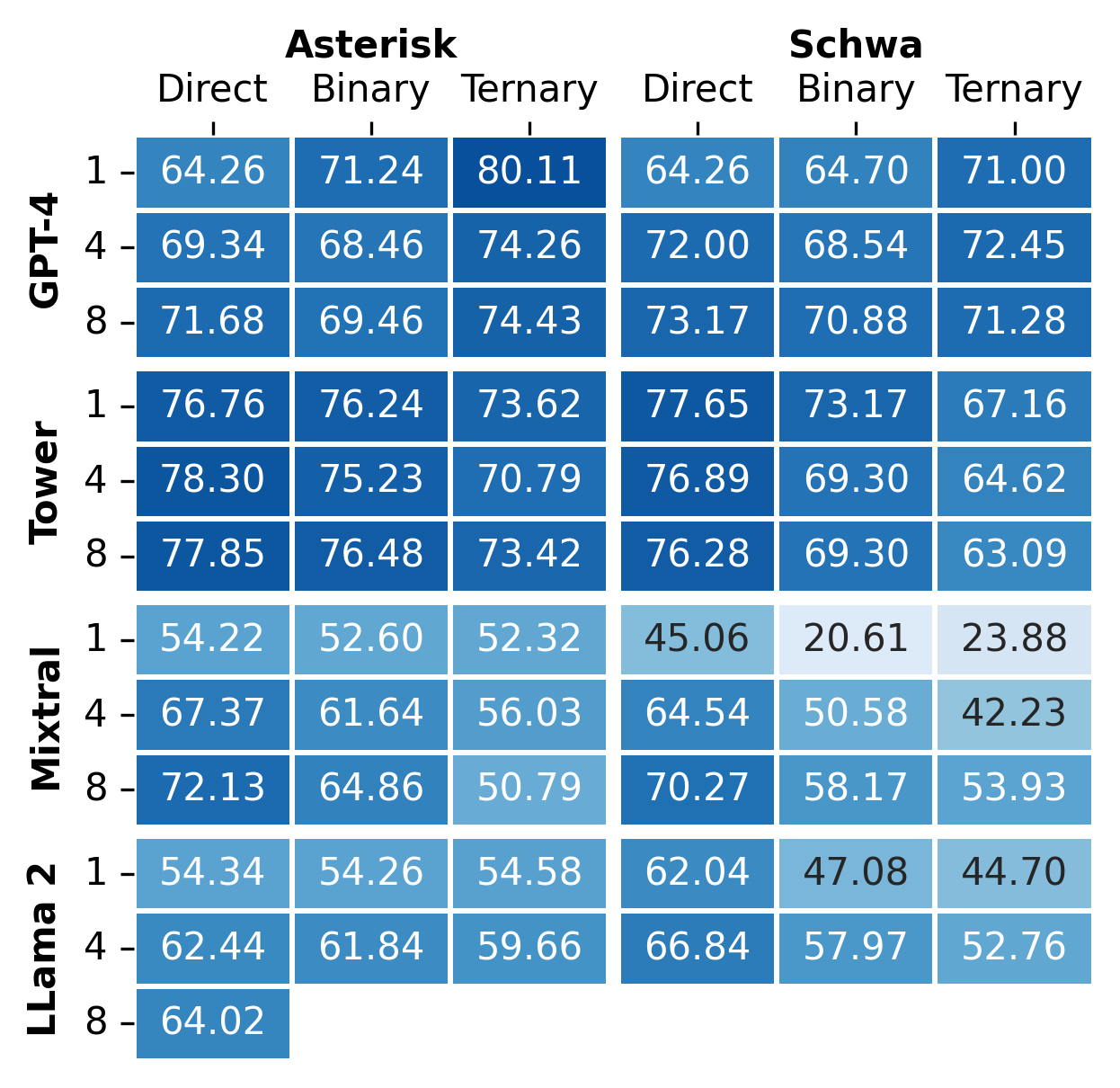}  
    \caption{\textbf{Coverage} percentage scores.}
    \label{fig:cov}
\end{subfigure}
\begin{subfigure}{.49\textwidth}
    \centering
    \includegraphics[width=.99\linewidth]{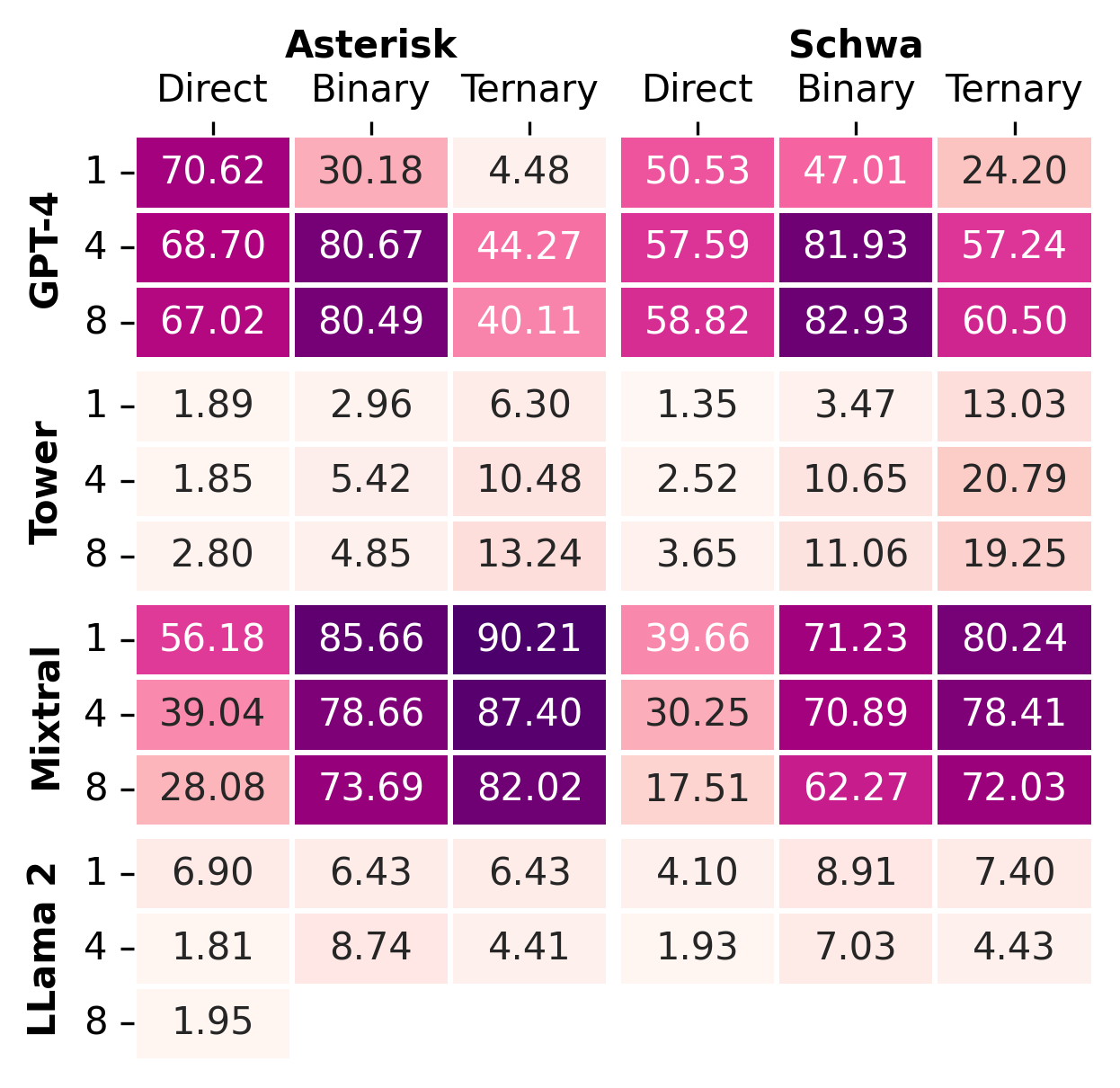}  
    \caption{\textbf{Accuracy} percentage scores.}
    \label{fig:acc}
\end{subfigure}
    \caption{Coverage and accuracy results in the few-shot settings. Darker shades indicate better performance.}
       %scores for all few-shots experiments settings. Higher scores, signalled by darker cell backgrounds, are better.}
    \label{fig:coverage-accuracy}
\end{figure*}

\subsection{Zero-shot results}
\label{sec:results-zero-shot}
The results of our zero-shot experiments are reported in 
%Table \ref{tab:results-zero-shot}. Here, we find two main kinds of behavior. 
Table \ref{tab:results-zero-shot}, which unveils very different model behaviors.
On the one hand, GPT-4 and Mixtral achieve significantly higher accuracy scores compared to the other two models,  
with GPT doubling Mixtral's performance. The accuracy scores 
indicate
%unveil
that, out of the matched terms, GPT correctly generated 74.63\% Asterisk neomorphemes and 60.19\% Schwa neomorphemes, with Mixtral reaching 37.92\% and 27.79\% respectively.
Accounting for coverage, the gap widens further, with GPT's coverage-weighted accuracy amounting to more than three times that of Mixtral (42.60 and 28.24, vs 13.35 and 8.35). Both models tend to produce considerable amounts of mis-generations, which are most often higher than the respective coverage scores. This implies that \textbf{GPT and Mixtral generate plenty of neomorphemes, but they use them incorrectly for the most part}. 
%Regardless, both models perform better with the Asterisk paradigm rather than the Schwa: with the first paradigm they achieve higher accuracy and lower mis-generation scores compared to the latter.
Regardless, according to all the metrics, both models perform better with the Asterisk paradigm rather than the Schwa, %possibly due to the use by the latter of distinct singular and plural forms, which is a further complication of the task.
possibly due to the latter's use of distinct singular (\textschwa) and plural (\textrevepsilon) forms, adding a further challenge to the task.

On the other hand, \textbf{LLama 2 and Tower severely %underperform
under-generate neomorphemes}, regardless of the paradigm.
More specifically, LLama's near zero accuracy scores (0.57 and 0.35) paired with its low mis-generation scores (16.70 and 12.79) indicate that LLama 2 rarely generates neomorphemes and, when it does, it uses them inaccurately. 
Finally, Tower's high coverage scores (77.57 and 77.25) combined with the rest of the metrics, all of which report 0 scores, indicate that 
%Tower
the model
produces fluent, gendered outputs and never generates neomorphemes in the zero-shot setting.  
%This is possibly
%\mn{This is likely}
This can be
due to the fact that in Tower-instruct's fine-tuning data set, TowerBlocks,\footnote{\url{https://huggingface.co/datasets/Unbabel/TowerBlocks-v0.1/}} 
%the characters we used in the neomorpheme paradigms are essentially absent: there are three occurrences of `\textschwa' in English segments and no occurrence of `\textrevepsilon' and `*' altogether.
our neomorpheme characters are practically absent (3 occurrences of `\textschwa' in English segments, and no occurrences at all of `\textrevepsilon' and `*').
%\mn{chiusa sul fatto che nonpossiamo trarre conclusion definitive ulterior perche' non sapiamo se * e schwa compaiono tra i dati di training di gpt/mixtral}
%
%
%
%\ap{However, as the development data of the other \mn{two} models is not publicly available, we cannot investigate this hypothesis any further and we cannot draw conclusions.}
However, since the development data of the other two models is not publicly available, we cannot further investigate this hypothesis %to 
and
draw definitive conclusions.

\subsection{Few-shots experiments}
\label{sec:results-few-shots}
\setlength{\tabcolsep}{3.5pt}
%%\looseness=-1 
For the 
%few-shots
few-shot
experiments we report each of the four metrics separately. We do not report all LLama 2 scores because in some cases, namely all the 8-shots settings, the model struggled to reproduce the format described in §\ref{sec:experimental-settings}. In such instances, LLama 2 failed to insert the angle brackets or the labels we included in our prompts, and its outputs contained too many hallucinations to be automatically post-processed and evaluated. As the model did not seem to yield better performance 
%nor to
or exhibit
interesting phenomena in 
%8-shots 
8-shot settings, the 
%cost of manually extracting
additional effort required  
%for extracting the output translation from 
to process
its unpredictable outputs was unjustified. 
%Thus,
Therefore,
we only 
%evaluated and reported 
report
the scores of one of the 8-shots settings outputs, namely the Asterisk, 
%8-shots, Direct prompt setting.
Direct prompt setting.

% \begin{figure*}[ht]
%     \centering
% \begin{subfigure}{0.49\textwidth}
%     \centering
%     \includegraphics[width=.99\linewidth]{figures/coverage.png}  
%     \caption{\textbf{Coverage} percentage scores.}
%     \label{fig:cov}
% \end{subfigure}
% \begin{subfigure}{.49\textwidth}
%     \centering
%     \includegraphics[width=.99\linewidth]{figures/accuracy.png}  
%     \caption{\textbf{Accuracy} percentage scores.}
%     \label{fig:acc}
% \end{subfigure}
%     \caption{Coverage and accuracy results in the few-shot settings. Darker shades indicate better performance.}
%        %scores for all few-shots experiments settings. Higher scores, signalled by darker cell backgrounds, are better.}
%     \label{fig:coverage-accuracy}
% \end{figure*}

%%\looseness=-1 
\subsubsection{Coverage and accuracy}
\label{sec:coverage-and-accuracy}
The coverage and accuracy scores are reported in Figure \ref{fig:coverage-accuracy}.
%\textbf{\bs{Looking}} at \textbf{coverage (\ref{fig:cov})} first, we find that in the few-shots settings, on average, the models either significantly improve their scores (GPT-4 and Mixtral) or perform similarly to the zero-shot setting (LLama 2 and Tower).
%We also note that for Mixtral and LLama 2 a higher number of demonstrations generally results in a higher coverage score, whereas Tower \bs{surprisingly} exhibits an opposite behavior. 
%
%\textbf{Looking at \textbf{coverage (\ref{fig:cov})} first, we note that it generally improved compared to the zero-shot results. We also notice a trend in Mixtral and LLama's coverage, which improves at higher numbers of demonstrations.}
\textbf{Looking at coverage (\ref{fig:cov}), we observe that few-shot prompting generally leads to improvements compared to the zero-shot results. 
%We also notice a trend in Mixtral and LLama's coverage, which improves at higher numbers of demonstrations.
Also, for Mixtral and LLama,
%\mn{the gains} 
the scores
increase at higher numbers of demonstrations.}
As for the prompts, the Direct %prompt 
format generally produces higher coverage scores, with only GPT performing better with the Ternary format. 
% Interestingly, the neomorpheme paradigm has an impact on coverage, as we see consistently higher scores with the Asterisk paradigm rather than the Schwa. This could be the result of models producing more correct generations with the Asterisk neomorpheme or more mis-generations with the Schwa paradigm, or both.
% Also in this case, but for all models, 
Interestingly,
the neomorpheme paradigm has an impact on coverage, as we see generally higher scores with the Asterisk paradigm compared to the Schwa. As discussed in \S\ref{sec:misgen}, this can be ascribed to the models' tendency to produce more mis-generations with the latter% paradigm.
.

%\looseness=-1 
% However, %\textbf{\bs{increases in coverage could be due to models generating neomorphemes correctly and in higher quantities as well as them producing more, fluent gendered forms. }}
% as coverage is only informative of the proportion of annotated terms the models generated, we look at
% \textbf{accuracy (\ref{fig:acc})} to understand how many of those words include neomorphemes.
%Here we first find that all models improve their performance in at least one setting, compared to the zero-shot experiments. 
Coverage, however, is only informative of the proportion of annotated terms the models generated and disregards how many of those words include neomorphemes.
\textbf{Looking at \textbf{accuracy (\ref{fig:acc})}  we 
%first find 
find
that all models improve their performance in at least one setting} compared to the zero-shot experiments, 
%\ap{consistently with findings reported by Hossein et al.~\shortcite{hossain-etal-2023-misgendered} in a mask filling task.
%
%\ap{consistently with findings of Hossein et al.~\shortcite{hossain-etal-2023-misgendered} \mn{about the effectiveness of in-context learning.}}
confirming the %usefulness 
benefits
of in-context learning for generative tasks involving neologistic expressions \cite{hossain-etal-2023-misgendered}.
Mixtral and GPT are confirmed as the models which produce the highest rates of correct neomorphemes, with the first topping at 90.21 and the latter at 82.93 accuracy. On the contrary, Tower and LLama 2 are unfit for this task despite their improvements, as their scores remain low.

\begin{figure}[t]
\centering
\includegraphics[scale=0.731]{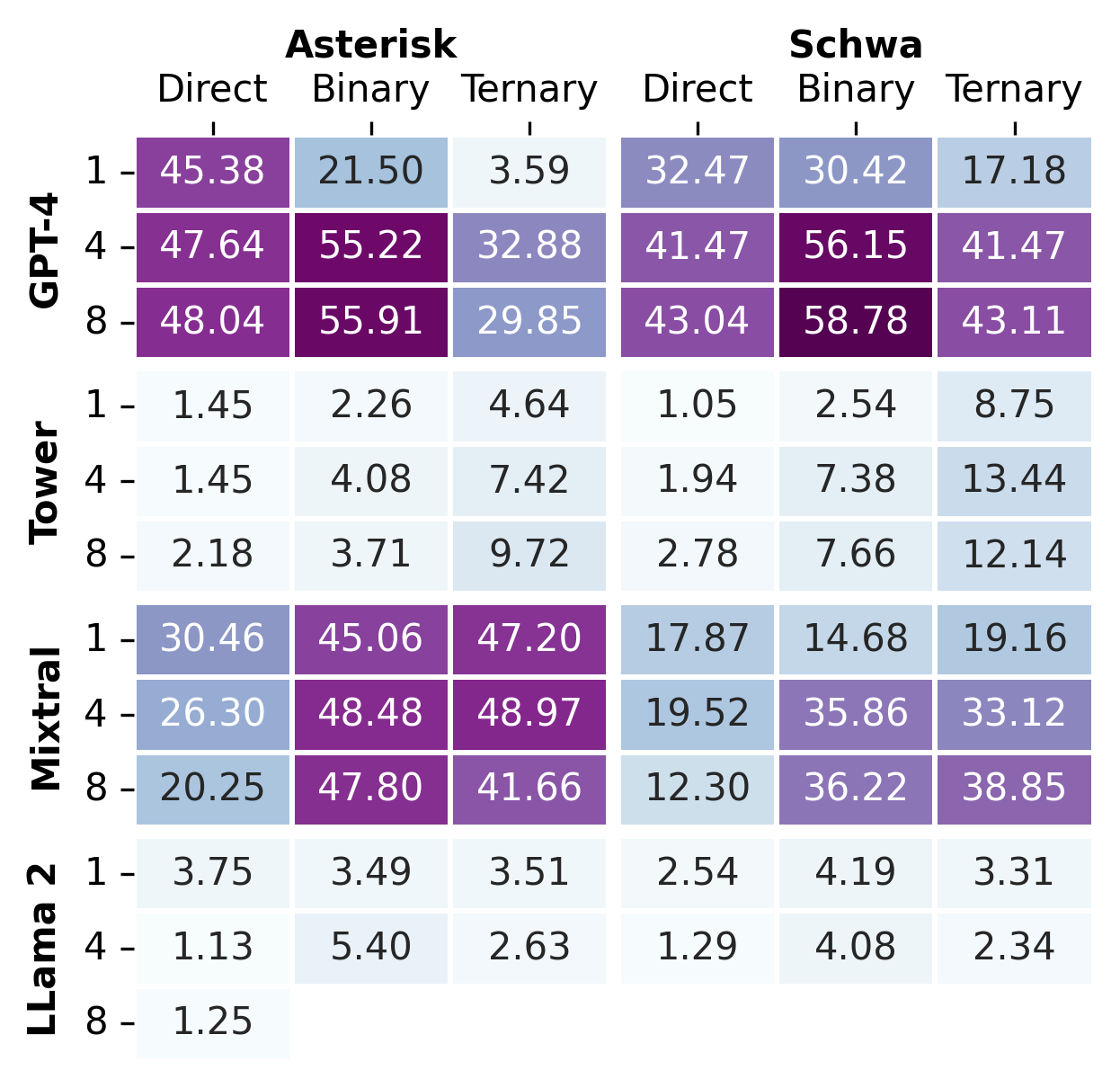}
\caption{\textbf{Coverage-weighted accuracy} percentage scores 
%of all models, prompts, and demonstration settings 
for the few-shot settings. Darker shades indicate better performance.
%Higher scores (darker cells), are better.
}
\label{fig:cov-weighted-accuracy}
\end{figure}

\begin{figure}[t]
\centering
\includegraphics[scale=0.731]{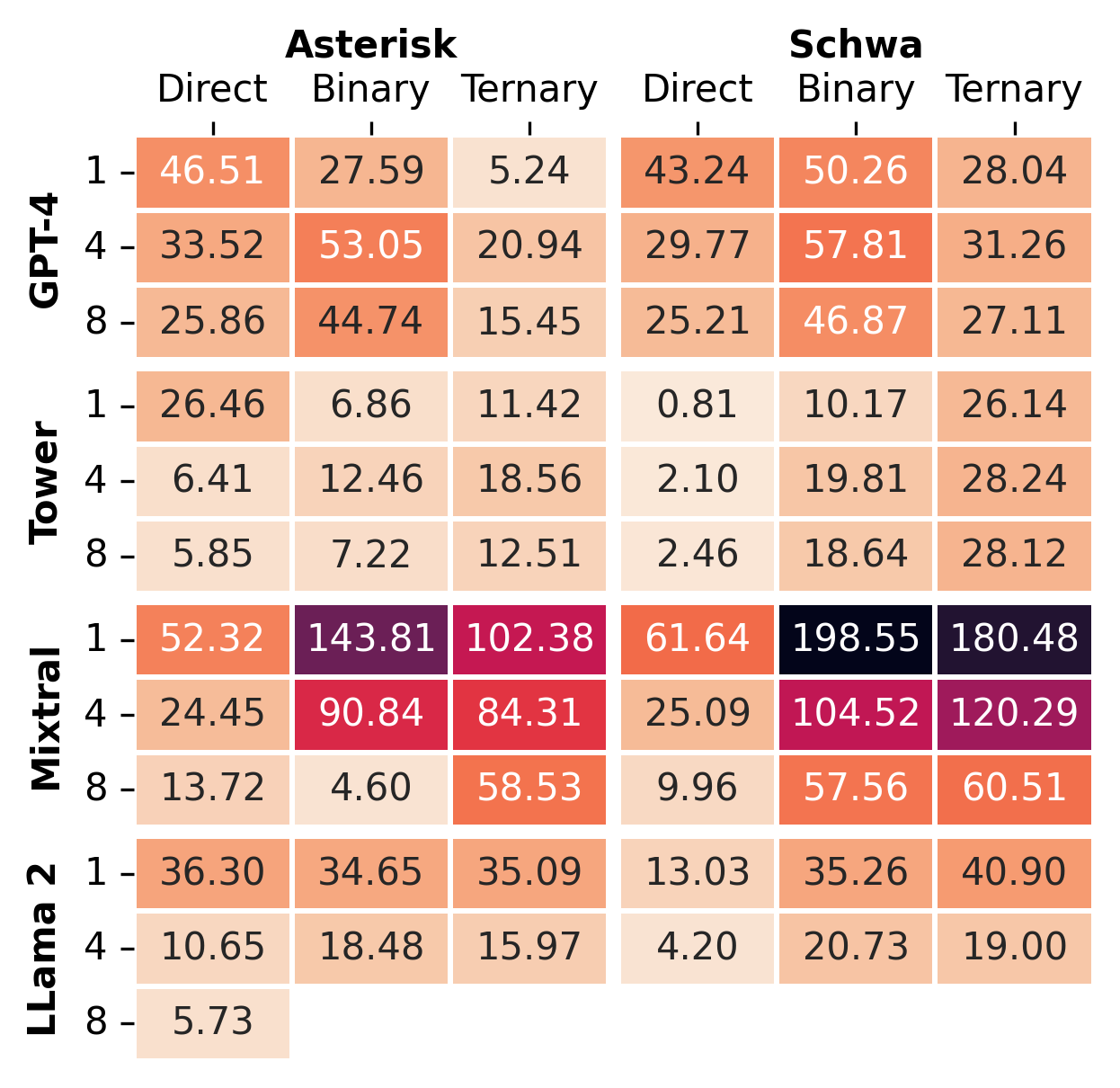}
\caption{\textbf{Mis-generation} percentage scores for the few-shot settings. % 
Higher scores (darker shades) indicate worse performance.
%cells), 
%are worse.
}
\label{fig:misgen}
\end{figure}

Surprisingly, 
%\textbf{a higher number of demonstrations does not guarantee a better accuracy.}
\textbf{a greater number of demonstrations does not necessarily lead to higher accuracy.}
While coverage generally increased with more demonstrations, 
%for accuracy this is only true for GPT and Tower, meaning that they generate more neomorphemes and do so in a more correct way.
this trend generally holds true for accuracy only %in the case of 
for 
GPT and Tower, indicating that they generate more neomorphemes and do so more accurately.
On the contrary, 
%LLama 2 and Mixtral's accuracy 
the accuracy of LLama 2 and Mixtral
significantly decreases with more 
%demonstrations, which, paired with their rising coverage, indicates 
demonstrations. Paired with their rising coverage, this indicates
that they produce fewer neomorphemes and more gendered terms. Both behaviors may 
%be results of 
result from
systems better 
%`understanding' 
modeling the task 
%given
with
more demonstrations, as LLama's and Mixtral's higher accuracy and lower coverage in the 1 shot settings may be due to fortuitous correct generations of neomorphemes in a context where they over-generate them. 
We discuss this aspect %further 
below, looking at mis-generation (§\ref{sec:misgen}).

As for the neomorpheme paradigms, Mixtral and Tower perform better with the Asterisk and the Schwa %paradigms 
respectively, as in the zero-shot experiments. GPT does not seem to be consistently affected by the neomorpheme %paradigm here, and
paradigm.
LLama presents negligible differences between the paradigms as well. 
%\textbf{\bs{Differences in performance between the two paradigms, as well as the lack thereof, could depend on the representation of the specific characters we used in our neomorpheme paradigms in each model's training set. We can only investigate this aspect with respect to Tower, as the training sets of the other models are not publicly available. As mentioned in §\ref{sec:results-zero-shot}, `\textschwa' is the only character we used as a neomorpheme appearing in Tower's training set, though with only three occurrences. This could explain Tower's better performance with the Schwa paradigm, as the model may assign higher probabilities to a character that appeared in its training set, as is the case of `\textschwa', rather than to ones that did not, as is the case of `\textrevepsilon' and `*'. A manual inspection of Tower's best setting results (Schwa, Ternary, 4 shots) supports this hypothesis, as out of the 333 correct neomorpheme generations it produced 309 included the neomorpheme `\textschwa' and only 24 included `\textrevepsilon', despite the two being represented in the prompt quite equally (the 4 shots Schwa Ternary prompt includes 9 instances of `\textschwa' and 10 instances of `\textrevepsilon').}}
The ability to more correctly generate one neomorpheme over another is possibly due to %the 
models' robustness to likely unseen grammatical paradigms and to the representations of the specific characters used as neomorphemes in each model's training data. Unfortunately, we cannot investigate this aspect as such data is not publicly available, with the exception of Tower's fine-tuning data set, as mentioned in §\ref{sec:results-zero-shot}.

\subsubsection{Coverage-weighted accuracy}
%%\looseness=-1 
%\textbf{\bs{COMMENTO NEL LATEX}}

%\bs{Secondo me qui va chiarito il senso della sezione: aiutare a bilanciare il giudizio di quale possa essere la combinazione di prompt/modello e shot migliore. Pesando sia coverage che accuracy (both how accurately a model generates neomorphemes and the proportion of the entire dataset covered by the evaluation script): la combinanzione sara' il modello che meglio gestisce i neomorfemi al netto della sua capacita' traduzione/generare le parole corrette. Alla luce di questo, confermiamo mixtral e gpt best models. Focusing on prompts, per Mixtral, i best results sono nel blocco binario/ternario con + shots, sia per tower sebenne negligible. GPT invece chiaramente nel blocco binario 4/8 shots. Ignora gli altri due che tanto fanno schifo.}
%
%%\looseness=-1 
To %\pending{properly} 
compare models' overall performance in gender-inclusive 
%MT properly, 
MT,
we look at coverage-weighted accuracy in Figure \ref{fig:cov-weighted-accuracy}. This metric offers a comprehensive view of model performance in each setting, allowing for a fair comparison of different systems in light of both coverage and accuracy. 

%%\looseness=-1 
We first find that \textbf{all models improve their performance in the few-shots experiments}. The upside offered by 
%ICL
in-context learning
is notable, and there is arguably room for improvement at higher numbers of demonstrations.
% GPT and Mixtral are confirmed as the best models. Moreover, here the gap between GPT and Mixtral 
% %shortens 
% narrows
% significantly, with respect to the zero-shot experiments. 
GPT and Mixtral are confirmed as the best models, and the gap between them narrows significantly with respect to the zero-shot experiments.
In the best configurations, GPT scores
58.78 (Schwa, Binary, 8 shots)
and Mixtral scores 48.97 (Asterisk, Ternary, 4 shots). 
%Thus, GPT-4 emerges as an upper bound for gender-inclusive MT, whereas Mixtral stands out as the best open model. 
%%%%% SACRIFICABILE PER SPAZIO 
Generally, GPT performs better with the Binary prompt and 
4 or 8 
%\ap{4/8}
shots, whereas Mixtral achieves its best results in the Binary/Ternary, 4/8 shots region, especially with the Asterisk paradigm.
%%%%% %%%%% %%%%% %%%%% %%%%%
As for the other two models, despite the very low scores Tower generally outperforms the ten times bigger LLama 2, but both come across as unfit for this task.

%Overall, all models improve their performance significantly in the few-shots experiments. The upside offered by ICL is notable, and there is arguably room for improvement at higher numbers of demonstrations, \textbf{as all models appear to perform progressively better given more demonstrations, with the exception of LLama 2}. As for the prompts, there is no single format granting better performance from all models. In general, the Direct prompt does not seem to conceptualize the task as well as the more complex formats, but the preference for a Binary or a Ternary format appears to be model specific.

% \begin{figure}[t]
% \centering
% \includegraphics[scale=0.731]{figures/cov-weighted_acc.png}
% \caption{\textbf{Coverage-weighted accuracy} percentage scores 
% %of all models, prompts, and demonstration settings 
% for the few-shot settings. Darker shades indicate better performance.
% %Higher scores (darker cells), are better.
% }
% \label{fig:cov-weighted-accuracy}
% \end{figure}

% \begin{figure}[t]
% \centering
% \includegraphics[scale=0.731]{figures/misgen.png}
% \caption{\textbf{Misgeneration} percentage scores for the few-shot settings. % 
% Higher scores (darker shades) indicate worse performance.
% %cells), 
% %are worse.
% }
% \label{fig:misgen}
% \end{figure}

\subsubsection{Mis-generation}
\label{sec:misgen}

So far, % our \bs{few-shot discussion} %evaluation 
the discussion of our few-shots experiments
has %only 
focused on the correct generation of neomorphemes when referring to human entities, thus only on relevant phenomena %that 
we  
annotated in our test set. 
%%% SPOSTATO SU
%However, under realistic conditions, it might be the case that models overgeneralize and generate neomorphemes also for unrelated words, compromising the intelligibility of the rest of the sentence. To account for such a scenario and thus better inform the capabilities of the investigated models, as a final analysis we also estimate the number of \textit{mis-generations} they produce, i.e. words that include neomorphemes which are not annotated in our test set 
%%% %%% %%% %%%
To better investigate models' behavior, we look at mis-generation, i.e. inappropriate neomorphemes generations, as well (Figure \ref{fig:misgen}). %(see §\ref{sec:evaluation}). 
%Aside from assessing whether the models use neomorphemes correctly, we also investigate whether they use them incorrectly, looking at the mis-generation scores in Figure \ref{fig:misgen}. As discussed in §\ref{sec:evaluation}, we consider \textit{mis-generations} those words which include neomorphemes and do not match \textsc{Neo-GATE}'s annotation. 
%\pending{Examples of mis-generations found in our experiments are reported in Table \ref{tab:examples-misgen}}%, in Appendix \ref{sec:appendix-misgen}.

We first note that \textbf{Mixtral stands out as the model producing the most mis-generations}, especially in the 1 and 4 shots, Binary and Ternary region. 
%1/4 shot, Binary/Ternary region. 
Table \ref{tab:example-mixtral-misgen} reports examples of mis-generation from Mixtral's %Schwa, 1 shot Binary prompt 
outputs.

\begin{table}[h]
    \centering
    \small
    \begin{tabular}{ll}
        \textbf{Source} & \begin{tabular}[c]{@{}p{5.5cm}@{}} I hope the shaman can help us. \end{tabular} \\
        \textbf{Annotation} & \begin{tabular}[c]{@{}p{5.5cm}@{}} lo la l\textschwa; sciamano sciamana sciaman\textschwa; \end{tabular} \\
        \textbf{Output} & \begin{tabular}[c]{@{}p{5.5cm}@{}} \textbf{\ul{Sper\textschwa}} che \ul{l\textschwa} \ul{sciaman\textschwa} possa aiutarci. \end{tabular} \\
        \midrule
        \textbf{Source} & \begin{tabular}[c]{@{}p{5.5cm}@{}} They asked everyone to remain silent. \end{tabular} \\
        \textbf{Annotation} & \begin{tabular}[c]{@{}p{5.5cm}@{}} tutti tutte tutt\textrevepsilon; \end{tabular} \\
        \textbf{Output} & \begin{tabular}[c]{@{}p{5.5cm}@{}} Hanno chiesto a \textbf{\ul{t\textrevepsilon}} di \textbf{\ul{rimaner\textschwa}} in \textbf{\ul{silenzi\textschwa}}. \end{tabular}
    
    \end{tabular}

    \caption{Examples of mis-generation found in Mixtral's Schwa, 1 shot, Binary prompt outputs. Words containing neomorphemes are underlined, mis-generations are %also 
    in bold.}
    \label{tab:example-mixtral-misgen}
\end{table}

As hypothesized in §\ref{sec:coverage-and-accuracy}, in these settings Mixtral over-generates neomorphemes, resulting in both correct generations and mis-generations. This behavior is reflected in the high accuracy and low coverage: by over-generating neomorphemes, Mixtral produces fewer gendered words -- which would contribute to coverage -- and many words that are either \textit{a)} correct or \textit{b)} mis-generations. 
With more task demonstrations, Mixtral generates significantly fewer mis-generations, and while its accuracy decreases the coverage improves, meaning that it produces better formed outputs.
Mixtral's example testifies to how the mis-generation metric complements the analysis of models' behavior, as it sheds light on unwanted phenomena related to neomorphemes usage, which coverage and accuracy alone (or combined) cannot signal.

Similarly to Mixtral, LLama~2 produces more mis-generations given fewer demonstrations and progressively mitigates this behavior when given more.
With an opposite trend, GPT generates fewer mis-generations, and 
Tower even less. However, the best performing settings of both models 
are also the ones
% \mn{those}
in which they produce the most mis-generations. Hence, future work should focus on improving the ratio of 
correctly generated neomorphemes over the total 
%\mn{correct generations over the total}
neomorphemes
generated by these models.

\section{Conclusions}
\label{sec:conclusions}

We discussed a neologistic approach to gender-inclusive 
%MT, 
machine translation,
an underexplored area constrained by the lack of publicly available dedicated data.
Our first contribution, the release of the \textsc{Neo-GATE} benchmark, allowed us to give a first fundamental impulse to research in this direction. As a second contribution, we explored the possibility of performing gender-inclusive 
%en$\rightarrow$it 
translation from English to Italian with four popular 
%LLMs 
Large Language Models:
three open models -- Mixtral, Tower, and LLama 2 -- and 
%one commercial 
a commercial one
-- GPT-4.
Our comparisons across different prompting settings reveal that GPT-4 and Mixtral generally exhibit promising results when properly prompted, while LLama 2 and Tower are unfit for the task. More specifically, models' understanding of the task is significantly influenced by prompt complexity, the number of demonstrations, and the specific characters employed as neomorphemes (possibly depending on the representation of those characters in each model's training data). 

While our investigation suggests LLMs’ potential for neologistic gender-inclusive MT, there remains %\mn{scope} for enhancing 
room for improving
their accuracy. 
\textsc{Neo-GATE} and the analyses presented herein lay the groundwork for rising to the challenge and for future research on gender-inclusive MT tailored to existing neologistic paradigms, and those that may emerge in this new and evolving landscape.

\section{Acknowledgements}
We acknowledge the support  of the PNRR project FAIR -  Future AI Research (PE00000013),  under the NRRP MUR program funded by the NextGenerationEU.

\bibliography{eamt24}
\bibliographystyle{eamt24}

\appendix

\section{Tagset and annotation}
\label{sec:appendix-tagset}

Table \ref{tab:tagset} reports the complete tagset used in \textsc{Neo-GATE}, as well as the tagset mappings for the Schwa and the Asterisk paradigms.

\section{Function words anchoring}
\label{sec:function-words-annotation}
We include an additional %annotation
information for function words which maps them to an anchor, in respect to which they are expected to be correctly positioned. This check allows for a more precise evaluation of function words, as it ensures that the evaluation is performed on the appropriate function word, and not on other ones which may occur in the sentence.

An anchor consists in the longest possible sub-word common to the masculine, feminine, and tagged content word which the function word is syntactically linked to. 
Looking at Table \ref{tab:anchors-examples}, the first Annotation reports an example of anchor for a function word: `student=1'. It indicates that the sub-word sequence `student' is the anchor for the function word forms `il la l*', meaning that if one of the three forms is found it will only be evaluated if the anchor is found immediately after it (i.e., at a distance of 1 word).
Similarly, the second annotation of the table reports anchor annotations for two function words. The first, `amic=2' indicates that if one of the three forms `i le l*' is found, it will only be evaluated if the anchor `amic' is found at a distance of two words. The second anchor annotation `amic=1' maps the function word forms `tuoi tue tu*' to the same anchor `amic', which should be positioned one word after.

We did not include anchor annotations in the main body to simplify the examples. However, all function words annotated in \textsc{Neo-GATE} are assigned with anchors, including the ones reported in the examples throughout the paper.

\setlength{\tabcolsep}{2pt}
\begin{table*}[t]
\small
\centering
\begin{tabular}{llllll}
\toprule
\textbf{TAG}& \textbf{Description} & \textbf{Masculine}&  \textbf{Feminine}&  \textbf{Asterisk} & \textbf{Schwa} \\
\midrule
\textless{}ENDS\textgreater{}& \begin{tabular}[c]{@{}p{6.5cm}@{}} inflectional morpheme (word ending), singular \end{tabular}  & o, e, tore& a, essa, trice& *  & \textschwa~ \\
\textless{}ENDP\textgreater{}& \begin{tabular}[c]{@{}p{6.5cm}@{}} inflectional morpheme (word ending), plural \end{tabular} & i, tori& e, esse, trici&  * & \textrevepsilon~  \\
\textless{}DARTS\textgreater{}& \begin{tabular}[c]{@{}p{6.5cm}@{}} definite article, singular \end{tabular}  & il, lo, l'& la, l'&  l* & l\textschwa~ \\
\textless{}DARTP\textgreater{}&  \begin{tabular}[c]{@{}p{6.5cm}@{}} definite article, plural \end{tabular} &i, gli& le&  l* & l\textrevepsilon~\\
\textless{}IART\textgreater{}& \begin{tabular}[c]{@{}p{6.5cm}@{}} indefinite article \end{tabular} & uno, un& una, un'& un* & un\textschwa~\\

%\textless{}PARTS\textgreater{}& del, dello, dell'& della, dell'& de\textschwa~& de* \\
\textless{}PARTP\textgreater{}& \begin{tabular}[c]{@{}p{6.5cm}@{}} partitive article, plural \end{tabular} & dei, degli& delle& de*& de\textrevepsilon~ \\
 
%\textless{}PRON\textgreater{}& lui& lei& l\textschwa~i& l*i \\
\textless{}PREPdiS\textgreater{}& \begin{tabular}[c]{@{}p{6.5cm}@{}} articulated preposition with root `di', singular\end{tabular}  & del, dello, dell'& della, dell' & dell* & dell\textschwa~     \\
\textless{}PREPdiP\textgreater{}& \begin{tabular}[c]{@{}p{6.5cm}@{}} articulated preposition with root `di', plural\end{tabular} & dei, degli& delle& dell*  & dell\textrevepsilon~    \\
\textless{}PREPaS\textgreater{}& \begin{tabular}[c]{@{}p{6.5cm}@{}} articulated preposition with root `a', singular\end{tabular} & al, allo, all'& alla, all'& all*& all\textschwa~\\
\textless{}PREPaP\textgreater{}& \begin{tabular}[c]{@{}p{6.5cm}@{}} articulated preposition with root `a', plural\end{tabular} & agli, ai& alle& all* & all\textrevepsilon~\\
\textless{}PREPdaS\textgreater{}& \begin{tabular}[c]{@{}p{6.5cm}@{}} articulated preposition with root `da', singular\end{tabular}  & dal, dallo, dall'& dalla, dall'& dall* &dall\textschwa~     \\
\textless{}PREPdaP\textgreater{}& \begin{tabular}[c]{@{}p{6.5cm}@{}} articulated preposition with root `da', plural\end{tabular}  & dagli& dalle& dall* & dall\textrevepsilon~     \\
%\textless{}PREPinS\textgreater{}& nel, nello, nell'& nella, nell'& nell\textschwa~& nell*      \\
\textless{}PREPinP\textgreater{}& \begin{tabular}[c]{@{}p{6.5cm}@{}} articulated preposition with root `in', plural\end{tabular}  & negli& nelle& nell*  & nell\textrevepsilon~    \\
\textless{}PREPsuS\textgreater{}& \begin{tabular}[c]{@{}p{6.5cm}@{}} articulated preposition with root `su', singular\end{tabular}  & sul, sullo, sull'& sulla, sull'& sull* & sull\textschwa~     \\
\textless{}PREPsuP\textgreater{}& \begin{tabular}[c]{@{}p{6.5cm}@{}} articulated preposition with root `su', plural\end{tabular} & sugli& sulle& sull* & sull\textrevepsilon~     \\
\textless{}DADJquelS\textgreater{}& \begin{tabular}[c]{@{}p{6.5cm}@{}} demonstrative adjective (far), singular\end{tabular} & quel, quello, quell'& quella, quell'& quell* & quell\textschwa~    \\
\textless{}DADJquelP\textgreater{}& \begin{tabular}[c]{@{}p{6.5cm}@{}} demonstrative adjective (far), plural\end{tabular}  & quegli& quelle& quell* & quell\textrevepsilon~    \\
\textless{}DADJquestS\textgreater{}&  \begin{tabular}[c]{@{}p{6.5cm}@{}} demonstrative adjective (near), singular\end{tabular} & questo, quest'& questa, quest'& quest*& quest\textschwa~     \\
\textless{}DADJquestP\textgreater{}& \begin{tabular}[c]{@{}p{6.5cm}@{}} demonstrative adjective (near), plural\end{tabular}  & questi& queste& quest* & quest\textrevepsilon~    \\
\textless{}POSS1S\textgreater{}& \begin{tabular}[c]{@{}p{6.5cm}@{}} possessive adjective, 1st person singular, singular\end{tabular}  & mio& mia& mi* & mi\textschwa~ \\
\textless{}POSS1P\textgreater{}& \begin{tabular}[c]{@{}p{6.5cm}@{}} possessive adjective, 1st person singular, plural\end{tabular}  & miei& mie& mi*& mi\textrevepsilon~ \\
\textless{}POSS2S\textgreater{}& \begin{tabular}[c]{@{}p{6.5cm}@{}} possessive adjective, 2nd person singular, singular\end{tabular} & tuo& tua& tu* & tu\textschwa~ \\
\textless{}POSS2P\textgreater{}& \begin{tabular}[c]{@{}p{6.5cm}@{}} possessive adjective, 2nd person singular, plural\end{tabular} & tuoi& tue& tu* & tu\textrevepsilon~ \\
\textless{}POSS3S\textgreater{}&  \begin{tabular}[c]{@{}p{6.5cm}@{}} possessive adjective, 3rd person singular, singular\end{tabular} & suo& sua& su* & su\textschwa~ \\
\textless{}POSS3P\textgreater{}& \begin{tabular}[c]{@{}p{6.5cm}@{}} possessive adjective, 3rd person singular, plural\end{tabular} & suoi& sue& su* & su\textrevepsilon~ \\
\textless{}POSS4S\textgreater{}& \begin{tabular}[c]{@{}p{6.5cm}@{}} possessive adjective, 1st person plural, singular\end{tabular}  & nostro& nostra& nostr* & nostr\textschwa~     \\
\textless{}POSS4P\textgreater{}& \begin{tabular}[c]{@{}p{6.5cm}@{}} possessive adjective, 1st person plural, plural\end{tabular} & nostri& nostre& nostr* & nostr\textrevepsilon~     \\

%\textless{}POSS5S\textgreater{}& vostro& vostra& vostr\textschwa~& vostr*     \\

%\textless{}POSS5P\textgreater{}& vostri& vostre& vostr\textrevepsilon~& vostr*     \\
\textless{}PRONDOBJS\textgreater{}& \begin{tabular}[c]{@{}p{6.5cm}@{}} direct object pronoun, singular \end{tabular} & lo& la& l* & l\textschwa~  \\
\textless{}PRONDOBJP\textgreater{}& \begin{tabular}[c]{@{}p{6.5cm}@{}} direct object pronoun, plural \end{tabular} & li& le& l*& l\textrevepsilon~  \\

%\textless{}PRONIOBJS\textgreater{}& gli& le& l\textschwa~& l*  \\

%\textless{}PRONIOBJP\textgreater{}& gli& gli& l\textrevepsilon~& l*   \\
\bottomrule
\end{tabular}
\caption{The full tagset used in \textsc{Neo-GATE} and the tagset mappings to the Italian gendered forms and the desired forms in the Asterisk and Schwa nomorpheme paradigms.}
\label{tab:tagset}
\end{table*}

\section{Translation experiments prompt}
\label{sec:appendix-mt-prompt}

Table \ref{tab:translation-prompt} reports the prompt we used %in the general translation preliminary experiments 
to assess the general translation quality of the systems, as
discussed in §\ref{sec:models}. We include three demonstrations taken from FLORES' dev set, so as to provide the LLMs with an interaction structure to reproduce. This facilitates the process of filtering out extra comments and hallucination produced by the models, and extract the output translation.

\begin{table*}
    \centering
    \small
    \begin{tabular}{ll}
    \toprule
        \textbf{Source} & \begin{tabular}[c]{@{}p{11.5cm}@{}}  The student was worried about going off topic. \end{tabular} \\
        \textbf{Tagged reference} & \begin{tabular}[c]{@{}p{11.5cm}@{}} L* student* era preoccupat* di andare fuori tema. \end{tabular} \\
        \textbf{Annotation} & \begin{tabular}[c]{@{}p{11.5cm}@{}} lo la l* \textbf{student=1}; studente studentessa student*; preoccupato preoccupata preoccupat*; \end{tabular} \\
        \midrule
        \textbf{Source} & \begin{tabular}[c]{@{}p{11.5cm}@{}} Come out to the balcony and let your friends see you. \end{tabular} \\
        \textbf{Tagged reference} & \begin{tabular}[c]{@{}p{11.5cm}@{}} Esci sul balcone e lascia che l* tu* amic* ti vedano. \end{tabular} \\
        \textbf{Annotation} & \begin{tabular}[c]{@{}p{11.5cm}@{}} i le l* \textbf{amic=2}; tuoi tue tu* \textbf{amic=1}; amici amiche amic*; \end{tabular} \\
        \midrule
    \end{tabular}
    \caption{Examples of \textsc{Neo-GATE} annotations including anchors. The Tagged references and the Annotations are adapted to the Asterisk paradigm, the anchors are in bold.}
    \label{tab:anchors-examples}
\end{table*}

\setlength{\tabcolsep}{4pt}
\begin{table*}[ht]
    \centering
    \small
    \begin{tabular}{ll}
    \toprule
         \textbf{Role} & \textbf{Content} \\
        \midrule
         \texttt{user} & \begin{tabular}[c]{@{}p{13cm}@{}} Translate the following English sentence into Italian: \textless{}Construction is ongoing for five new skyscrapers at the site, with a transportation center and memorial park in the middle.\textgreater{}\end{tabular}\\
         \texttt{assistant} &  \begin{tabular}[c]{@{}p{13cm}@{}} \textless{}Nel sito sono in corso i lavori di costruzione di cinque nuovi grattacieli, con un centro trasporti e un parco alla memoria nel mezzo.\textgreater{} \end{tabular} \\
         \texttt{user}& \begin{tabular}[c]{@{}p{13cm}@{}} \textless{}The game is based on the Second Battle of Fallujah, a vicious battle between American and Iraqi forces.\textgreater{} \end{tabular} \\
         \texttt{assistant}& \begin{tabular}[c]{@{}p{13cm}@{}} \textless{}Il videogioco simula la seconda battaglia di Fallujah, un conflitto cruento tra le forze armate americane e quelle irachene.\textgreater{} \end{tabular} \\
         \texttt{user}& \begin{tabular}[c]{@{}p{13cm}@{}} \textless{}Born in the Croatian capital, Zagreb, Bobek gained fame while playing for Partizan Belgrade.\textgreater{} \end{tabular} \\
         \texttt{assistant}& \begin{tabular}[c]{@{}p{13cm}@{}} \textless{}Nato nella capitale croata Zagabria, Bobek è diventato famoso giocando nel Partizan Belgrado.\textgreater{} \end{tabular} \\
         \texttt{user}& \begin{tabular}[c]{@{}p{13cm}@{}} \textless{}\{\texttt{input sentence}\}\textgreater{} \end{tabular} \\
         \bottomrule
    \end{tabular}
    \caption{The 3 shots prompt used in the general translation preliminary experiments.}
    \label{tab:translation-prompt}
\end{table*}

\end{document}